\documentclass[authoryear]{elsarticle}
\usepackage[utf8]{inputenc}

\usepackage{amsmath, amssymb, amsfonts, bm}
\usepackage{graphicx}
\usepackage{textcomp}
\usepackage{xcolor}
\usepackage{booktabs, multirow, makecell, colortbl}
\usepackage[font=scriptsize]{caption}
\usepackage{comment}
\usepackage{hyperref}
\hypersetup{
    urlcolor  = black,
	colorlinks = true,
	citecolor = blue, % citation color
	linkcolor = blue % internal links color
}
\usepackage[ruled]{algorithm2e}
\usepackage{parskip} % remove indent and set space among paragraphs
\usepackage[a4paper, total={6in, 8in}]{geometry}
\usepackage{natbib}
\setcitestyle{authoryear,open={(},close={)}}
\usepackage{subcaption}
\usepackage{setspace}

\usepackage{tikz}
\usetikzlibrary{shapes, decorations, arrows, calc, arrows.meta, fit, positioning}
\tikzset{
    -Latex,auto,node distance =1 cm and 1 cm,semithick,
    state/.style ={ellipse, draw, minimum width = 0.7 cm},
    point/.style = {circle, draw, inner sep=0.04cm,fill,node contents={}},
    bidirected/.style={Latex-Latex,dashed},
    el/.style = {inner sep=2pt, align=left, sloped}
}

\DeclareMathOperator{\pa}{\mathbf{pa}}
\newcommand{\w}[1]{\widetilde{#1}}
\newcommand{\ra}[1]{\renewcommand{\arraystretch}{#1}}
\newcommand{\rowcol}{\rowcolor{gray!20}} %

\begin{document}

\title{Counterfactual Explanations as Interventions in Latent Space}

\author[1]{Riccardo Crupi\fnref{dr}}
\ead{riccardo.crupi@intesasanpaolo.com}

\author[1]{Alessandro Castelnovo\fnref{dr}%\corref{cor}
}
\ead{alessandro.castelnovo@intesasanpaolo.com}

\author[1]{Daniele Regoli\fnref{dr}}
\ead{daniele.regoli@intesasanpaolo.com}

\author[2]{Beatriz San Miguel Gonzalez}
\ead{beatriz.sanmiguelgonzalez@fujitsu.com}

\address[1]{Data Science and Artificial Intelligence, Intesa Sanpaolo S.p.A., Torino, Italy}
\address[2]{Fujitsu Laboratories of Europe, Madrid, Spain}

\fntext[dr]{The views and opinions expressed are those of the authors and do not necessarily reflect the views of Intesa Sanpaolo, its affiliates or its employees.}

\begin{abstract}
Explainable Artificial Intelligence (XAI) is a set of techniques that allows the understanding of both technical and non-technical aspects of Artificial Intelligence (AI) systems. XAI is crucial to help satisfying the increasingly important demand of \emph{trustworthy} Artificial Intelligence, characterized by fundamental characteristics such as respect of human autonomy, prevention of harm, transparency, accountability, etc. Within XAI techniques, counterfactual explanations aim to provide to end users a set of features (and their corresponding values) that need to be changed in order to achieve a desired outcome. Current approaches rarely take into account the feasibility of actions needed to achieve the proposed explanations, and in particular they fall short of considering the causal impact of such actions.
In this paper, we present Counterfactual Explanations as Interventions in Latent Space (CEILS), a methodology to generate counterfactual explanations capturing by design the underlying causal relations from the data, and at the same time to provide feasible recommendations to reach the proposed profile. Moreover, our methodology has the advantage that it can be set on top of existing counterfactuals generator algorithms, thus minimising the complexity of imposing additional causal constrains. We demonstrate the effectiveness of our approach with a set of different experiments using synthetic and real datasets (including a proprietary dataset of the financial domain). 
\end{abstract}


\maketitle

\onehalfspacing

% ================================================================================================= %
\section{Introduction}

The use and importance of Artificial Intelligence (AI) systems and, in particular, Machine Learning (ML) models, has increased in many industrial sectors (i.e. finance, healthcare, hiring, transportation, etc.), with the purpose, among others, to automate decision-making processes. The majority of these systems have a direct or indirect impact on people's life. Besides the benefits, many ethical concerns have recently emerged with the widespread use of such systems: bias amplification, data privacy, lack of transparency, human oversight, accountability, etc.~\citep{pekka2018european}. This is putting increasing pressure on developers and service providers to supply explanations of the models and in particular of their outcomes. The European Commission has recently published a proposal for what is going to be the first attempt ever to insert AI systems and their use in a coherent legal framework~\citep{EUproposal}: the proposal devotes a significant attention to the importance of \emph{transparency} of AI systems.

To enhance transparency and trust in AI systems, Explainable Artificial Intelligence (XAI) has become increasingly important~\citep{adadi2018peeking}. In a nutshell, XAI aims to provide information to explain and justify automated results and thus to give the tools to understand the AI system behaviour. The ultimate goals of XAI are many, among which the prevention of possible harms to end users and also the possibility of gaining useful insights to improve the system itself. Therefore, XAI involves the ability to explain both the technical aspects, pertaining to modeling, and the related human decisions. This entails to consider different target audiences of the proposed explanations, such as data scientists, developers, executives, regulatory entities and end users affected by decisions, among others~\citep{arrieta2020explainable}. Against this background, a growing number of methods and approaches have appeared, depending on the type of problem faced and the stakeholder considered  (see, e.g., \cite{adadi2018peeking, arrieta2020explainable}).

Explainabilty in Machine Learning (ML) is usually faced either by employing simple and thus intrinsically understandable models, such as linear or logistic regressions, rule-based methods, decision trees, etc., or by using appropriate tools to generate explanations on trained models.
These tools are usually categorized as local or global methods, and as model-specific or model-agnostic solutions~\citep{molnar2019, guidotti2018survey}.
While local methods aim to provide explanations for a single instance and its given outcome, global methods aims to explain the overall behavior of the model. An example of a global method is that of approximating a black-box model with an intrinsically explainable one.
Model-specific approaches are tools that can be applied only to some classes of models (e.g. looking at weights of a linear regression), while model-agnostic methods can be used to explain any black-box model.
LIME~\citep{lime} and SHAP~\citep{shap} are two of the most popular model-agnostic local explainers, and they are part of the broader family of methods based on feature contribution, together with other well-known approaches such as Partial Dependence Plots~\citep{pdp}, Accumulated Local Effects~\citep{ale}, Individual Conditional Expectation~\citep{ice}.
Other methods are based on local rule extraction, such as Anchors~\citep{anchors} and LORE~\citep{lore}.
These model-agnostic approaches provide explanations by trying to locally approximate the black-box model. One the other hand, example-based methods explain local instances by computing other points in the feature space --- the \emph{examples} --- with some desired characteristics, such as representing the typical point belonging to some class (prototypes and criticism~\citep{kim2016examples, gurumoorthy2019efficient}), or being similar to the original point but with enough changes to be given a different outcome (adversarial examples --- see \cite{molnar2019} --- and counterfactual explanations~\citep{wachter2017counterfactual}). 

%However, the majority of these developments are focused on technical profiles rather than end users affected by the decisions~\citep{bhatt2020explainable, jin2021euca}. From the perspective of end users, explanations are useful to understand and thus possibly contest the automated decisions, and also to take actions in order to receive a different result~\citep{wachter2017counterfactual}. 

Counterfactual explanations, first proposed by~\cite{wachter2017counterfactual}, are becoming one of the most popular approaches to explainability in ML within technical, legal and business circles~\citep{barocas2020hidden}. They are local example-based and mostly model-agnostic explanations that construct a set of statements to communicate to end users what could be changed from an original input to receive an alternative decision. Unlike other explainability techniques, this approach imposes no constraints on model complexity, avoids the disclosure of technical details (protecting trade secrets), provides precise suggestions for achieving a desired outcome and appears to produce explanations that comply with requirements of note-worthy governmental initiatives~\citep{barocas2020hidden}. 

The volume of research on counterfactual explanations is growing and different solutions have been proposed (refer to \cite{verma2020counterfactual, stepin2021survey} for surveys on counterfactual explanations methods). While these efforts are significant, generally they fall short of generating feasible actions that end users should carry out~\citep{barocas2020hidden}, which is the focus of this work. Obviously, the concept of causality is a key element if we want to find counterfactual explanations that guide end users to act and not only understand the output of a model~\citep{chari2020directions}. Causal methods can effectively represent cause-effect relationships among variables, thus going in the direction of disentangling the causal effects on the entire system due to direct changes in some of the variables.

In this paper we present CEILS: Counterfactual Explanations as Interventions in Latent Space, a method to generate counterfactual explanations capturing by design the underlying causal relations from the data. 
This methodology is based on the idea of employing existing counterfactual explanations generators on a latent space that represent the residual degrees of freedom once the causal structure of the problem is taken into account.
We demonstrate the effectiveness of our approach with a set of different experiments using synthetic and real datasets (including a proprietary dataset of the financial domain). We evaluate the explanations using a large set of metrics, trying to quantify and pinpoint the key aspects of our proposal. This evaluation is a useful precursor to user studies, where interactions with users and feedback are employed to guide towards the best explanations.

The paper is organised as follows. Section~\ref{sec:background} covers the existing background and related work on counterfactual explanations. Section~\ref{sec:solution} details our proposed methodology, whose main advantages with respect to prior works are highlighted in section~\ref{sec:recourse}. Section~\ref{sec:experiments} is devoted to experiment results to evaluate our method, including a detailed definition of the metrics and a discussion of the major findings. Finally, section~\ref{sec:conclusions} concludes the paper by summarizing the proposed method and presenting its main limitations and possible directions of improvement.

% ================================================================================================= %
\section{Related work \label{sec:background}}

\subsection{Related concepts}

Several governmental initiatives towards explainable AI, such as the General Data Protection Regulation (GDPR) in the European Union~\citep{gdpr} and the Defence Advanced Research Projects Agency (DARPA) XAI program of the United States~\citep{darpa}, as well as the already mentioned European Commission proposal for legislation on AI systems~\citep{EUproposal}, endeavour to promote the creation of explanations that can be understood and appropriately trusted by end users. With the goal of approaching user-centric explanations in AI, researchers can use findings from previous research on social science, wherein contrastive and counterfactual explanations are claimed to be inherent to human cognition~\citep{miller2019explanation}.

In the field of XAI, there seem to be an overlap in the concepts of \emph{contrastive} and \emph{counterfactual} explanations~\citep{stepin2021survey}. An explanation is \emph{contrastive} when it does not describe the reason for an event to happen (\textit{“Why P?”}), but seek for the reason of an event to occur \emph{relative to another} that did not (\textit{“Why P rather than Q?”})~\citep{miller2019explanation}. Counterfactual explanations are defined as a set of statements constructed to communicate what could be changed in the original profile to get a \emph{different} outcome by the decision-making process~\citep{wachter2017counterfactual}. Therefore, counterfactual explanations are normally considered contrastive by nature and give a source of valuable complementary information \citep{byrne2019counterfactuals}. Indeed, people usually do not ask why a certain prediction was made, but why this prediction was made \emph{instead of another prediction}: therefore, one of the usual requirements for a ``good'' explanation is precisely to be contrastive~\citep{lipton1990contrastive, molnar2019}. Notice that counterfactual explanations have the additional characteristic of representing a conditional clause  (\textit{"If X were to occur, then Y would (or might) occur"})~\citep{stepin2021survey}, thus adding a ``causality layer'' on the contrastive statement. Indeed both the work of \cite{karimi2020algorithmic} and our proposed methodology present a bridge between ``counterfactuals'' as intended by causal inference frameworks~\citep{pearl2016causal, spirtes2000causation} and ``counterfactual explanations'' that are usually \emph{not} embedded in formal causal theory frameworks.

Counterfactual explanations are strictly connected to, but different from, \emph{algorithmic recourse}. While the former, as the name suggests, provides an \emph{explanation} of a specific model outcome (by means of showing a scenario as close as possible to the original but reaching a different outcome), the latter provides \emph{recommendations} of what action to undertake in order to gain a different outcome. In layman’s terms, counterfactual explanations inform an
individual where they need to get to, but not how to get there \citep{karimi2021algorithmic}.
Rephrasing \cite{karimi2021survey}, both can be cast in a counterfactual form by asking the following questions: 
\begin{itemize}
    \item \emph{explanation}: what profile would have led to receiving a different outcome?
    \item \emph{recourse}: what actions would have led me to reach such profile?
\end{itemize} 
Algorithmic recourse refers, in fact, to the set of actions that an individual should perform in order to reach the desired outcome~\citep{joshi2019towards, venkatasubramanian2020philosophical}. Notice that the second question somehow incorporates the first one. In other terms, algorithmic recourse is a broad concept, which contains both the counterfactual explanation and the recommendations on how to reach it. 

To address the challenge of algorithmic recourse, it is important do distinguish the variables in terms of their level of ``actionability'' \citep{karimi2020algorithmic}: there are variables that cannot change (e.g. race, sex, date of birth), variables that can change but cannot be directly controlled by the individual (e.g. credit score), and variables that can --- at least in principle --- be directly acted upon (e.g. bank balance, income, education).  

The aforementioned difference between explanations and recourse may seem only a matter of terminology, and indeed in the majority of the literature on counterfactual explanations it is understood that, given a counterfactual observation, it is straightforward to find the set of actions necessary to reach it by simply taking the difference of the two feature vectors~\citep{barocas2020hidden}. But this is true only under very stringent assumptions, that are outlined in~\cite{karimi2020algorithmic2, karimi2021survey} and that will be made clearer in section~\ref{sec:recourse}. 

\subsection{Generation of explanations}

Since the first proposal of counterfactual explanations by~\cite{wachter2017counterfactual}, a large body of research concerning different algorithms and techniques to generate contrastive and counterfactual explanations have been conducted~\citep{verma2020counterfactual, stepin2021survey}. Most of generation techniques relies on establishing an optimization problem to find the \emph{nearest counterfactual} in the space of features, with respect to the observation to be explained~\citep{wachter2017counterfactual, mace, mohammadi2020scaling}. The metric used to define the distance to be minimized is sometimes referred to as \emph{proximity}. 
Moreover, several additional proposals have been put forward to achieve desirable explanatory properties, such as keeping a low number of feature changes (\emph{sparsity})~\citep{dice} or possibly producing more than one counterfactual explanation per each observation, as \emph{diverse} as possible among each other~\citep{dice, mace}. \cite{alibi} propose the use of \emph{prototypes}~\citep{kim2016examples, gurumoorthy2019efficient} to guide the optimization process, with a twofold goal: to find counterfactuals that are ``as close as possible'' to the distribution of the observed dataset; to speed up the optimization search. \cite{dhurandhar2018explanations} propose the use of autoencoders trained on the given data to provide explanation that are near the data manifold.

Some works have put forward proposals to embed causality aspects into the counterfactual generation process. For example, \cite{mahajan2019preserving} introduce a ``causal proximity'' loss term by which the counterfactual explanation is ``pushed'' towards regions in which the underlying causality among features is preserved. \cite{karimi2020algorithmic} build on the optimization framework described in \cite{mace} and introduce causality by computing counterfactuals through abduction-action-prediction steps as prescribed by~\cite{pearl2016causal}. \cite{karimi2020algorithmic2} focus on the case in which only limited causal knowledge is available, thus propose probabilistic approaches (via Gaussian processes or conditional average treatment effect) that allow to relax the assumption of having access to full structural equations.

The counterfactual generation process is usually expressed as an optimization problem, either constrained or unconstrained (see section \ref{sec:prob_setting} for details) that has been faced with various strategies: gradient-based methods~\citep{wachter2017counterfactual, alibi, dice}; genetic-based algorithms~\citep{certifai, lore}; graph-based shortest path algorithms~\citep{face}; by building on formal verification tools and satisfiability modulo theories (SMT) solvers \citep{mace}. %\cite{dandl2020multi} translate the counterfactual search into a multi-objective optimization problem.

Finally, the literature takes into account other aspects as well: \cite{mahajan2019preserving} report the evaluation of counterfactual explanations with respect to the \textit{computational efficiency} and the amount of time necessary to generate the explanations, which is indeed one of the challenges in this field~\citep{verma2020counterfactual}.  \cite{binns2018s} and \cite{fernandez2020explaining} evaluate counterfactual explanations in comparison with other XAI approaches, such as feature importance. \cite{miller2019explanation} reviews relevant papers from disciplines such as philosophy, cognitive science and social psychology, to draw some findings that can be applied to AI.

% ================================================================================================= %
\section{Methodology \label{sec:solution}}

\subsection{Problem setting}\label{sec:prob_setting}

Consider having a ML classifier $\mathcal{C}$ trying to estimate the relationship between a binary target random variable $Y \in \{0, 1\}$ and predictors $X = (X_1, \ldots, X_d)$. %\footnote{The predictors $X$ can be a selection of the total predictors for example the only one that cause $Y$.}
Typically, one has
\begin{equation}
    \hat{Y} = \mathcal{C}(x) = \bm{1}_{\{R(x) > t\}},
\label{eq:M}
\end{equation}
where $x\in\mathcal{X}$ is a specific realization of $X$ --- namely, an observation; $R(x)$ is usually referred to as \emph{score} and is learned to estimate $P(Y = 1\ \rvert\ X=x)$; while $t$ is the threshold above which we assign the positive outcome to the observation $x$.

Given an instance $x^0$ we want to find a \emph{counterfactual explanation} for $x^0$, i.e. a $x^{0, \text{cf}} \in \mathcal{X}$ such that $\mathcal{C}(x^{0, \text{cf}}) \neq \mathcal{C}(x^0)$. Of course, the simple requirement that $x^0$ and $x^{0, \text{cf}}$ have different outcomes based on classifier $\mathcal{C}$ (condition that is referred to as \emph{validity} of $x^{0, \text{cf}}$~\citep{dice, verma2020counterfactual}) is a necessary but not sufficient condition to provide ``good'' counterfactual explanations.

The general formulation can be written as follows~\citep{karimi2021survey}: 
\begin{equation}
    \left\{
    \begin{aligned}
    &x^{0, \text{cf}} = \arg\min_{x\in\mathcal{P}_\mathcal{X}}\ dist(x, x^0), \\
    &\mathcal{C}(x) \neq \mathcal{C}(x^0);
    \end{aligned}\right.
    \label{eq:cff}
\end{equation}
where $dist: \mathcal{X} \times \mathcal{X} \rightarrow \mathbb{R}^+$ is a suitable distance function over $\mathcal{X}$.
The solution of problem~\eqref{eq:cff} provides the \emph{nearest counterfactual explanation} relative to observation $x^0$.
The space $\mathcal{P}_\mathcal{X}\subseteq\mathcal{X}$ is the subset of feature space $\mathcal{X}$ containing \emph{plausible} counterfactuals, i.e. it embodies a set of requirements that $x^{0, \text{cf}}$ should have in order to represent a realistic set of features with respect to the distribution of training data.

Following~\cite{karimi2020algorithmic}, we shall distinguish between \emph{plausibility} and \emph{feasibility} constraints.
Plausibility constraints refer to all the requirements expressed in feature space that go in the direction of having counterfactual explanations realistic with respect to the observed distribution. Feasibility, on the other hand, refers to the fact that a specific counterfactual $x^{0, \text{cf}}$ is actually reachable with a set of actions from the original observation $x^0$. The following example will help clarifying the distinction. An individual who is denied a loan may receive a counterfactual explanation where the age is reduced. While this may be perfectly plausible in terms of observed distribution (namely, there are observations in line with the proposed counterfactual) this is definitely not feasible for that individual to reach the proposed counterfactual.
This distinction is valuable in particular when discussing \emph{recommendations} besides explanations, i.e. the algorithmic recourse problem (see section~\ref{sec:recourse}).

The optimization problem~\eqref{eq:cff} is often relaxed in terms of an unconstrained loss minimization problem (see e.g. \cite{wachter2017counterfactual, alibi, mace}):
\begin{equation}
    x^{0, \text{cf}} = \arg\min_{x \in \mathcal{X}}\left(L_y(x, x^0) + \lambda\ dist(x, x^0) + \sum_i \beta_i L_P^i(x, x^0)\right);
\label{eq:loss}
\end{equation}
where we have:
\begin{itemize}
    \item the $L_y$ term pushing the outcome $y$ corresponding to $x$ away from that of $x^0$, i.e. pushing towards the \emph{validity} of $x^{0, \text{cf}}$;
    \item the $dist$ term keeping $x$ close to $x^0$ in feature space (\emph{proximity}),
    \item the $L_P^i$ terms guiding the solution towards \emph{plausible} points in $\mathcal{X}$.
\end{itemize}
The parameters $\lambda, \{\beta_i\}_i$ control the relative importance of each term. 
As mentioned in the previous section, several proposal have been put forward for each of the terms in the loss \eqref{eq:loss}, and in particular for the plausibility terms, to have more realistic, and thus useful explanations to be given to end users. For example, \cite{alibi} propose to use a term that penalizes the distance between $x$ and the nearest prototype of the class other than $\mathcal{C}(x^0)$, while \cite{dhurandhar2018explanations} add a term penalizing the distance between $x$ and its reconstruction by an autoencoder trained on the given dataset.

We here restrict to the case in which, for a given instance $x^0$, a unique counterfactual $x^{0, \text{cf}}$ is found, but it is also reasonable to suggest multiple counterfactuals per observation, which can be done either by simply running the minimization problem \eqref{eq:loss} multiple times with different random seeds as in \cite{wachter2017counterfactual}, or changing the formulation \eqref{eq:loss} to account for direct minimization over multiple counterfactual candidates, possibly diverse from each other to capture different aspects of the overall explanation, see e.g. \cite{dice, mace}.

\subsection{Counterfactual explanations in latent space}

We propose an algorithmic approach that builds on an arbitrary counterfactual explanation optimizer (namely, a strategy for solving a specific formulation of problem~\eqref{eq:loss}) but is able to find counterfactuals taking into account the underlying causal structure by design. In brief, we propose to generate explanations and corresponding recommendations by searching for nearest counterfactuals not in feature space, but in a latent space representing the residual degrees of freedom once the causal structure of the problem at hand is taken into account.
This approach has the advantage of providing the end users with feasible actions to reach a desired outcome, and of doing so with --- roughly speaking --- a ``simple'' change of variables on top of existing methodologies.

In doing so, we make use of causal graphs and Structural Casual Models (SCM) (which we discuss in more detail in subsections~\ref{sec:graph} and \ref{sec:SE}, respectively). This is in line with what proposed in \cite{karimi2020algorithmic}, but expands it in that we provide a methodology to compute causality-based counterfactual explanations without the necessity of a closed form SCM, and that can be used on top of any counterfactual generator, such as the ones proposed in \cite{wachter2017counterfactual}, \cite{alibi} or \cite{dice}. Indeed, other approaches to incorporate causality in counterfactual explanations generators either rely on an \emph{ad hoc} constrained optimization strategy~\citep{karimi2020algorithmic}, or they face the problem by adding additional terms in the loss~\eqref{eq:loss}~\citep{mahajan2019preserving}.

The contribution of our proposal is twofold: providing a straightforward way to incorporate causality relationships into the generation of counterfactual explanations for an arbitrary choice of the baseline generator of explanations; providing, besides counterfactual explanations, causal-aware recommendations for algorithmic recourse.

In a nutshell, our proposal can be summed up as: 
\begin{enumerate}
    \item use the SCM to translate the problem from feature space to the space of exogenous and root variables, that we shall call \emph{latent space} hereafter,
    \item apply an arbitrary counterfactual explanation optimizer on latent space,
    \item translate counterfactuals back to the original feature space.
\end{enumerate}

\subsection{Causal graph}
\label{sec:graph}

Our solution requires to access to a predefined causal graph that encodes the causal relationships among the variables of the dataset. Modeling causal knowledge is complex and challenging since it requires an actual understanding of the relations, beyond statistical evidence. Different causal discovery algorithms have been proposed to identify causal relationships from observational data through automatic methods~\citep{glymour2019review}. For example, the Python \texttt{Causal Discovery ToolBox}~\citep{kalainathan2020causal} includes many existing causal modeling algorithms such as PC~\citep{spirtes2000causation}, Structural Agnostic Model (SAM)~\citep{sam},  Max-Min Parents and Children (MMPC)~\citep{tsamardinos2003time}, etc.; the Python library \texttt{CausalNex} implements the NOTEARS alogorithm by~\cite{zheng2018dags}; the R packages \texttt{pcalg}~\citep{kalisch2012causal}, \texttt{kpcalg}~\citep{kpcalg}, \texttt{bnlearn}~\citep{bnlearn} include a vast selection of causal influence algorithms. In general, it is important that domain experts validate the relations detected by the causal discovery routine, or include new ones when deemed necessary. Moreover, experimentation based causal inference is also possible in some specific circumstances, e.g. via randomized experiments, and also by a combination of the observational and experimental methodologies~\citep{mooij2020joint}.

As usual, we model the underlying causal relationships among features by means of a Directed Acyclic Graph (DAG) $G = (V, E)$, with $V$ set of vertices (or nodes) and $E$ set of directed edges (or links). Nodes of the graph $G$ are composed by the actual variables $X = (X_1, \ldots, X_d)$  used as predictors in the model. Moreover, we denote with $U$ exogenous variables, representing factors not accounted for by the features $X$, and with $Y$ the dependent variable to be predicted/estimated by means of $X$. In causal graph theory, edges represent not only conditional dependence relations, but are interpreted as the causal impact that the source variable has on the target variable. We refer to nodes with no parents in $G$ as \emph{root nodes}, namely 
\[
 \text{root nodes} = \left\{v \in G\ \rvert\ \pa(v) = \emptyset\right\}.
\]

A Structural Causal Model (SCM) is a a triplet $(X, U, F) $ where $F: \mathcal{U} \rightarrow \mathcal{X}$ is a set of functions mapping the exogenous (unobserved) variables to the endogenous (observed) ones: $X = F(U)$. These are called Structural Equations (SE) and, besides describing which variables causally impacts which (that is already encoded in the graph $G$), they also determine \emph{how} these relations work. Therefore, SCM prescriptions are much stronger than simply prescribing a DAG.

The dataset $D_n = \left\{(x^1, y^1), \ldots, (x^n,y^n)\right\}$ is composed by $n$ i.i.d. realizations of $(X, Y)$. Each $x_i$ is a $d$-dimensional vector, each component representing an observed feature. In the same fashion, $\left\{u^1, \ldots, u^n\right\}$ represent the realizations of the unobserved variables $U$.

\subsection{Structural Equations}\label{sec:SE}

Structural Equations are relations describing the precise functional form that links latent variables $U$ to observable ones $X$. Assuming an Additive Noise Model (ANM) we have the following:
\begin{equation}
    X_v = f_v(\pa(X_v)) + U_v,\quad v = 1, \ldots, d.
\label{eq:ANM}
\end{equation}

Besides this assumption, we also assume \emph{causal sufficiency}, i.e. that there are no counfounders unaccounted for in the specified DAG.

\begin{figure}
\begin{center}
\vspace{3ex}
\begin{tikzpicture}
    % nodes
    %% endogenous
    \node[state] (a) at (-2, 0) {age};
    \node[state] (g) at (-2, -2) {gender};
    \node[state] (am) at (2, 0) {amount};
    \node[state] (d) at (2, -2) {duration};
    %% exogenous
    \node[state, fill=gray!30] (Ua) at (-4, 0.5) {$U_1$};
    \node[state, fill=gray!30] (Ug) at (-4, -1.5) {$U_2$};
    \node[state, fill=gray!30] (Uam) at (4, 0.5) {$U_3$};
    \node[state, fill=gray!30] (Ud) at (4, -1.5) {$U_4$};
    % edges
    \path (a) edge (am);
    \path (g) edge (am);
    \path (am) edge (d);
    \path[dashed] (Ua) edge (a);
    \path[dashed] (Uam) edge (am);
    \path[dashed] (Ug) edge (g);
    \path[dashed] (Ud) edge (d);
\end{tikzpicture}
\end{center}
\caption{Causal graph for the German credit dataset}
\label{fig:german_graph}
\end{figure}
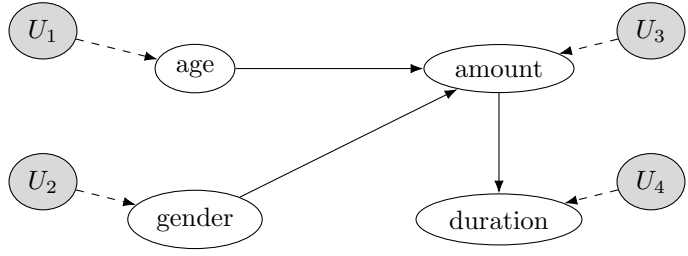

In figure~\ref{fig:german_graph} we see an example of DAG representing the causal relationships \cite{karimi2020algorithmic} in the German credit dataset~\citep{german}. Age and gender are root nodes, i.e. they are not caused by any other variable. $U_1, \ldots, U_4$ are the latent variables representing all the unobserved external causes. The unobserved causes for root notes, namely $U_1$ and $U_2$, are indeed redundant, namely $X_\text{root nodes}=U_\text{root nodes}$. The SE relative to figure~\ref{fig:german_graph} have the following form:
\begin{subequations}
\begin{align}
    &\text{age}  = U_1,\\
    &\text{gender}  = U_2,\\
    &\text{amount}  = f_3(\text{age}, \text{gender}) + U_3,\\
    &\text{duration}  = f_4(\text{amount}) + U_4.
\end{align}
\label{eq:SE_german}
\end{subequations}

In what follows, instead of specifying/assuming a precise form for each $f_v$ in equation \eqref{eq:ANM}, we are going to infer them from observations, namely from the collection $\{x^1, \ldots, x^n\}$. Specifically, in the spirit of~\cite{pawlowski2020deep}, we shall learn a regressor model $\mathcal{M}_v$ estimating $X_v$ from $\pa(X_v)$, and then compute the unobserved term as the residual:
\begin{equation}    
    \hat{U}_v = X_v - \hat{X_v} = X_v - \mathcal{M}_v(\pa(X_v)),\quad v = 1, \ldots, d; 
\label{eq:residual}
\end{equation}
which is the equivalent of $\hat{U} = \hat{F}^{-1}(X)$, where $\hat{F}$ are the SE estimated by data through $\mathcal{M}_v$ as discussed below.

For root nodes the model $\mathcal{M}_v$ is of course useless, and all the variability is encoded in the latent variable $U_v$.\footnote{It is just a matter of notation whether to introduce auxiliary latent variables also for root nodes, that we decide to follow for consistency of the formulation.}

Since for root nodes $r$ SE simply reduce to $F_r(U) = U_r$, once all models $\mathcal{M}_v$ are learned, following the causal flow in the DAG it is possible to recursively compute the actual function $F$ connecting $U$ to $X$ --- namely $X = F(U)$ --- by the following relation\footnote{With a slight abuse of notation we omit the estimation symbol $\hat{}$ over $F$ from here on.}:
\begin{equation}
    F_j(U) = \mathcal{M}_j\left(\left\{F_v(U)\right\}_{v\in\pa(X_j)}\right) + U_j,\quad \forall j\ \text{non-root}.
    \label{eq:ANM2}
\end{equation}

The procedure is summed up in Algorithm~\ref{alg:SE}.

\subsection{Model in the latent space}

Given the model $M$ with score function $R(x)$ as by \eqref{eq:M} and Structural Equations $X=F(U)$, it is straightforward to build their composition, effectively obtaining a model estimating $Y$ given $U$:
\begin{equation}
    \mathcal{C}_u(u) = (\mathcal{C} \circ F)(u) = \bm{1}_{\{R(F(u)) > t\}},
\end{equation}
where $R(F(u))$ is an estimate of $P(Y = 1\ \rvert\ U=u)$. Notice that the model $\mathcal{C}_u$ works precisely by following the causal flow of the underlying causal graph and its SCM. Namely, given some realization of the exogenous factors $U=u$, builds the corresponding values for the observed features by recursively applying \eqref{eq:ANM2}, i.e. by following the causal flow, and then predicts $Y$ by means of the initial model $\mathcal{C}$. Of course, we don't have access to exogenous variables, but this is not an issue when computing counterfactual explanations, since in that case we only need the value for $U = u^0$ corresponding to the instance that we need to explain ($x^0$) --- and possibly the values $\{u^1,\ldots, u^n\}$ corresponding to the training dataset $\mathcal{D}$ --- and these can be estimated by means of equation~\eqref{eq:residual}.

In other words, $\mathcal{C}_u$ is a model that 
takes in input $u$ in the residual (or latent) space, converts it in the original space $\mathcal{X}$ thanks to the SE $F(u)$ and then predicts its corresponding $Y$ with the model $\mathcal{C}$.

\subsection{Counterfactual generation}\label{sec:count_gen}

Once we have obtained the model $\mathcal{C}_u$ relating $U$ and $Y$, there are three more steps left to obtain the causal counterfactual explanation:
\begin{itemize}
    \item compute the latent variables $\{u^0, u^1, \ldots, u^n\}$,
    \item generate a counterfactual explanation $u^{0, \text{cf}} = u^0 + a$ of the observation $u^0$ for the model $\mathcal{C}_u$,
    \item given $u^{0, \text{cf}}$, compute the corresponding feature space counterfactual $x^{0, \text{cf}}=F(u^{0, \text{cf}})$.
\end{itemize}
The procedure is summed up in Algorithm~\ref{alg:CF}. 

Notice that these 3 steps reflect the usual steps of causal counterfactual computation~\citep{pearl2016causal}: abduction, intervention and prediction. 

\emph{Abduction} is the phase in which the possible events are restricted by the observation of the actual state of the world, namely $X=x^0$. In our framework, computing $U \ \rvert\ \{X = x^0\}$ is done via equation~\eqref{eq:residual}, i.e. as the residuals of regression models $\mathcal{M}_v$. 
Taking the example of the German dataset \eqref{eq:SE_german}, we would have:
\begin{subequations}
\begin{align}
    &u^0_1 = \text{age}^0,\\
    &u_2^0 = \text{gender}^0,\\
    &u_3^0 = f_3(\text{age}^0, \text{gender}^0) - \text{amount}^0 ,\\
    &u_4^0 = \text{duration}^0  - f_4(\text{amount}^0),
\end{align}
\label{eq:abduction_german}
\end{subequations}
where it is understood that the $f_i$'s need to be either known or estimated, e.g. as regressors $\mathcal{M}_i$.

\emph{Intervention} is the process of acting upon some variables and fixing it to some specific value. In our framework the actual intervention $a$ is computed --- via the minimization problem \eqref{eq:loss} applied to the model $\mathcal{C}_u$ for the observation $u^0$ --- as the minimal shift in latent space needed to reach a different outcome with respect to $\mathcal{C}_u$, namely $u^{0, \text{cf}} = u^0 + a$. Being it a shift on exogenous variables, this is actually a form of \emph{soft intervention} (see section~\ref{sec:recourse} for more details).

\emph{Prediction} step is the moment in which we compute the values of the observed variables $X$ given the latent $U=u^0$ and given the intervention $a$. In our framework, this is nothing but  $x^{0, \text{cf}}=F(a + u^0)$.
Taking again as reference the example of the German dataset \eqref{eq:SE_german}, \eqref{eq:abduction_german}, we have:
\begin{subequations}
\begin{align}
    &\text{age}^{0, \text{cf}} = u^0_1 +a_1,\\
    &\text{gender}^{0, \text{cf}} = u_2^0 + a_2,\\
    &\text{amount}^{0, \text{cf}} = f_3(\text{age}^{0, \text{cf}}, \text{gender}^{0, \text{cf}}) + u_3^0 + a_3 ,\\
    &\text{duration}^{0, \text{cf}} = f_4(\text{amount}^{0, \text{cf}}) + u_4^0 + a_4.
\end{align}
\label{eq:SE_prediction}
\end{subequations}

In the next section we discuss in more detail the role of actions and interventions in (causal) counterfactual explanations.

\begin{algorithm}
\footnotesize
\caption{Infer Structural Equations $F(U)$}
\label{alg:SE}

\SetKwInOut{Input}{input}\SetKwInOut{Output}{output}
\Input{Dataset of observed variables $\left\{x^1, \ldots,x^n\right\}$, causal graph $G = (V, E)$}
\Output{structural equations $F(U)$, $\{\mathcal{M}_v\}_{v \in V}$}

\BlankLine

\tcp{define auxiliary set of nodes for which Structural Equations are already computed, and initialize it with root nodes}
$V_\text{temp}=\left\{v\in V\ \rvert\  \pa(v)=\emptyset \right\}$\\
$F_v(U) = U_v$ for $v$ root nodes, i.e. such that $\pa(v) = \emptyset$\\
\tcp{loop until SE are computed for all nodes}
\While{$V_\text{temp}\neq V$ }{
    \tcp{loop over nodes in $G$}
    \For{$v$ in $V$}{
        \If{$\pa(v) \subseteq V_\text{temp}$  and $v \notin V_\text{temp}$}{
        train a model $\mathcal{M}_v$ estimating $X_v$ from $X_{\pa(v)}$ using data $\{x^1, \ldots, x^n\}$\\
        %\tcp{compute unobserved variables and store it}
        %$u^i_v = x^i_v - \mathcal{M}_v(x^i_{\pa(v)}),\ \forall i=1,\ldots,n$\\
        \tcp{build structural equations under additive model hypothesis}
        $F_v(U) = \mathcal{M}_v\left(\{F_j(U)\}_{j\in\pa(v)}\right)+U_v$\\
        \tcp{add node $v$ into the list of nodes for which SE are already computed}
        $V_\text{temp} = V_\text{temp} \cup \{v\}$\\
        }
    }
}
\KwResult{structural equations $F_v(U)$, $\mathcal{M}_v$ for all $v\in G$.}
\BlankLine

\end{algorithm}

\begin{algorithm}
\footnotesize
\caption{Train Model $\mathcal{C}$ and generate counterfactuals from residuals}
\label{alg:CF}

\SetKwInOut{Input}{input}\SetKwInOut{Output}{output}
\Input{dataset of observed variables $\left\{(x^1, y^1), \ldots,(x^n, y^n)\right\}$, factual observation $(x^0, y^0)$, structural equations $F(U)$, $\{\mathcal{M}\}_{v\in V}$}
\Output{counterfactual satisfying causal constrains $x^{0, \text{cf}}$ and action $a$}

\BlankLine

Train a classifier $\mathcal{C}$ with input dataset $\left\{x^1, \ldots,x^n\right\}$ and target $\left\{y^1, \ldots,y^n\right\}$.\\
\tcp{Build a model that estimates $y$'s from exogenous and root nodes}
$\mathcal{C}_u(U) = \mathcal{C}(F(U))$,  $\mathcal{C}_u: U \mapsto Y$\\
%$\mathcal{FC}(u)$ = $\mathcal{C}(F(u))$, such that final-model $: U \rightarrow y$\\
\tcp{Generate unobserved variables $\{u^0, u^1, \ldots, u^n\}$}
$V_\text{root}=\left\{v\in V\ \rvert\  \pa(v)=\emptyset \right\}$\\
$u^i_v = x^i_v,\ \forall v \in V_\text{root},\ \forall i=1,\ldots,n$\\
$u^i_v = x^i_v - \mathcal{M}_v\left(\{x^i_j\}_{j\in\pa(v)}\right),\ \forall v \notin V_\text{root},\ \forall i=0,\ldots,n$\\
\tcp{Generate explanation in latent space with a counterfactual generator CF}
%CF-model = CF.fit($\mathcal{C}_u, \left\{u^1, \ldots,u^n\right\}$)\\
$u^{0, \text{cf}}=\text{CF}\left(\mathcal{C}_u, \left\{(u^i, y^i)\right\}_{i=1}^n; u^0, y^0\right)$\\
\tcp{Compute action}
$a = u^{0, \text{cf}} - u^0$\\
\tcp{Compute explanation in feature space}
$x^{0, \text{cf}} = F(u^{0, \text{cf}})$\\

    \KwResult{action $a$, counterfactual explanation $x^{0, \text{cf}}$}
    \BlankLine

\end{algorithm}

% \section{Pseudocode (temporary version)}

% Functions to implement get-residuals and get-original-from-residuals.
% CF causal are done: 
% \begin{itemize}
% \item[1] Load data (U, V variables) and causal graph 
% \item[2] Compute the residual thanks to the structural equations in CFF class
% \item[3] Use root nodes (U) and residuals (eps) as the new dataset
% \item[4] Train a NN to estimate target (Y)
% \item[5] Use alibi to generate counterfactuals (U', eps')
% \item[6] Convert the counterfactuals (U', eps') to the orginal dataset (U', V')
% \end{itemize}

% Call $NN_1$ the NN that takes as input $X$ and output $Y$, $NN_2$ is the NN that takes as input $\varepsilon$ and output $Y$. $NN_1$ and $NN_2$ are equivalent since $f$ is deterministic and can be learned by $NN_2$ and then learn $NN_1$. 

% ================================================================================================= %
\section{Counterfactual explanations and recommended actions \label{sec:recourse}}

Counterfactual explanations defined as the nearest possible point to the original observation in feature space such that the model outcome is changed, may have problems in terms of realistic \emph{feasibility}. Indeed, while they may result useful in \emph{explaining} the reasons of an outcome for a specific instance --- by showing what features should have been different --- they may incur in problems when counterfactuals want to be seen as possible \emph{recommendations} on how to take \emph{actions} in order to get a different outcome. 
\citet{ustun2019actionable} introduces the framework of \emph{actionable recourse}, which tries precisely to fill the gap between counterfactual explanations and counterfactual recommendations.

Following \cite{karimi2020algorithmic, karimi2021survey}, it is useful to draw a line between the notions of \emph{plausibility} and \emph{feasibility} (or \emph{actionability}). As mentioned in section~\ref{sec:prob_setting}, we talk about plausibility constraints whenever we refer to conditions \emph{on the feature space} that pertain to have a realistic counterfactual explanation with respect to the training data distribution. Feasibility constraints, instead, are conditions on the \emph{actions} needed to reach some point in feature space. 
A couple of examples can clarify the apparent redundancy of these two concepts: a person with a low credit rating is unlikely to be granted a loan, thus an intuitive way to provide a counterfactual explanation is to suggest a profile with the same features but higher rating. One problem with this is that it may be an \emph{unrealistic} profile: since there may be other features correlated with rating, and thus the suggested profile may in fact be an outlier for the true distribution. Besides, there is another problem: how can the loan applicant \emph{reach} the suggested profile? Obviously, he cannot force it's rating to be higher: rating can change, but only as a consequence of the changes in other features, and these changes are not prescribed in the suggested profile. Therefore, the suggested profile is neither plausible nor feasible, for two different reasons. Take now the scenario in which a person is denied a loan because he is too old: in this case suggesting to be younger is of course useless since it is not feasible, but the resulting suggested profile would be, in general, perfectly plausible in terms of features.

Depending on the behavior with respect to actions, it is also useful to define (see~\cite{karimi2020algorithmic}):
\begin{itemize}
    \item \textbf{immutable} features, as those that cannot change in any way, neither for direct intervention nor for indirect consequences of changes in other variables;
    \item \textbf{mutable but non-actionable} features, as those that can change due to the changes in other connected features, but cannot be directly intervened upon (such as rating in the example above);
    \item \textbf{actionable} features, as the ones that can vary both due to indirect and direct interventions.
\end{itemize}

Recourse problem is defined by~\citet{ustun2019actionable} as a constrained optimization problem very similar to that of the nearest counterfactual~\eqref{eq:cff}:
\begin{equation}
    \left\{
    \begin{aligned}
    &a^* = \arg\min_{a\in\mathcal{\mathcal{F}_\mathcal{A}}}\ cost(a, x^0), \\
    &\mathcal{C}(x^0 + a) \neq \mathcal{C}(x^0),\\
    &x^0 + a \in \mathcal{P}_\mathcal{X};
    \end{aligned}\right.
    \label{eq:recourse}
\end{equation}
where $a^*$ is the ``cheapest'' \emph{action} --- in terms of the cost function $cost(a, x)$ --- that the individual identified by $x^0$ needs to perform in order to reach a different model outcome. The space $\mathcal{F}_\mathcal{A} \subseteq\mathcal{A}$ is that of all the actions $\mathcal{A}$ that are \emph{feasible}. 
It is straightforward to notice that, once defined $x = x^0 - a$ and $cost(a, x^0) = dist(x, x^0)$, the recourse problem~\eqref{eq:recourse} is equivalent to \eqref{eq:cff} a part for the explicit formulation of feasibility constraints.  

The problem with~\eqref{eq:recourse} is that it does not take into account the interdependence among variables and the fact that, in general, the change in a variable comes with changes in other variables as well. 
The natural framework to discuss how this interdependence impacts actions and counterfactual explanations is that of causal reasoning and in particular the notion of \emph{hard} and \emph{soft interventions}~\citep{eberhardt2007interventions} that we here try to summarize.
First of all, notice that the action computed as discussed in section~\ref{sec:count_gen}, namely:
\[
 a = u^{0, \text{cf}} - u^0,
\]
differs for root nodes and non-root nodes. For root nodes there is no difference between the latent variable $U_v$ and the feature $X_v$, thus in this case the action is simply the difference between the original and the counterfactual values of that feature. For non-root nodes, instead, the situation is different.
In general, we can express the relation between the real world features and the action with a straightforward application of SE, namely:
\begin{equation}
    x^{\text{cf}}_v - x_v = F(u^{\text{cf}}) - F(u) =  f_v(\pa(x^{\text{cf}}_v)) - f_v(\pa(x_v)) + a_v,
\end{equation}
i.e., the change in each feature is the sum of the change as a consequence of its parents' change and an explicit intervention. This is usually called a \emph{soft intervention}~\citep{eberhardt2007interventions}, since the explicit action $a_v$ is performed \emph{in addition} to the changes due to interventions on the ancestors. 
In contrast, an \emph{hard intervention} is identified by the following formula
\begin{equation}
    x^{\text{cf}}_v - x_v =  a_v,
\end{equation}
i.e. when acting on a variable $X_v$ we force it to assume the value $x_v + a_v$, somehow ``destroying'' or overriding any change due to its ancestors' changes. Therefore, our proposal is actually implementing by default soft interventions. Notice, however, that hard interventions are by far less interesting, and they are in any case much easier to account for: if we want to force an hard intervention on a variable then it is sufficient to cut out all the corresponding incoming edges in the causal graph. Indeed, as mentioned, for root nodes there is no distinction between hard and soft interventions. In other words, hard interventions are simply soft interventions on a modified casual graph.

In general, if we label with $\bm{N},\bm{I}$ the set of immutable and actionable features, respectively, we can write the intervention as follows:
\begin{equation}
    x^{\text{cf}}_v - x_v =    \left[f_v(\pa(x^{\text{cf}}_v)) - f_v(\pa(x_v))\right]\bm{1}_{X_v \notin \bm{N}} + a_v\bm{1}_{X_v\in\bm{I}}.
\end{equation}
At the practical level, imposing that a variable is mutable but non-actionable is straightforward in our implementation: it simply results in keeping the corresponding latent space variable fixed. 
On the other hand, imposing that a variable $A$ is immutable is straightforward when it is a root node, while if it has parents we need to consider carefully what \emph{immutable} means in the specific situation: if it means that $A$ must be kept fixed irrespective of the values of its parents, then it is equivalent to perform a null hard intervention, i.e. we need to change the graph by removing its incoming edges and at the same time freeze its latent variable $U_A$; if it means that a variable is fixed and that all the variables that can causally impact it should be fixed as well, then we need to treat as non-actionable both $A$ and all its ancestors up to the root level, i.e. we cannot change any of $A$ ancestors, or it would result in possible changes in $A$.
In the experiments we have done (see section~\ref{sec:experiments}) we consider immutable features only at the root level, thus the above distinction on the concept of immutability does not apply.

It is now clear that in the standard recourse optimization problem~\eqref{eq:recourse}, the action definition as \mbox{$x^{0, \text{cf}} = x^0 + a$} is equivalent to saying that all actionable variables are acted upon via hard interventions, i.e. each action is seen as enforcing a change in the feature overriding any other change due to variables interdependence. While this may be realistic in some specific cases, it cannot be taken as a paradigm for the general picture. Indeed, a combination of hard and soft interventions is very likely to be the most common situation.  

Moreover, even if we assume realistic that all actions are hard interventions, there still may be a causal flow impacting mutable but non-actionable variables. Therefore, neglecting completely the causal structure and sticking to the formulation~\eqref{eq:recourse} is equivalent to both assuming that all actions are hard interventions \emph{and} that there are no mutable but non-actionable variables. Indeed, without a causal structure, the distinction between immutable and non-actionable variables loses meaning.

\subsection{Computing \emph{ex post} actions of a given counterfactual explanation}
\label{sec:action_feature_space}

We have tried to clarify the reasons why computing algorithmic recourse without taking into account the causal structure of predictors results in a very specific and not much realistic form of interventions. We now try to address the following issue: instead of finding counterfactual explanations via a latent space representation and then computing actions as differences in the latent variables, why don't we find counterfactual explanations $x^*$ via ``standard'' algorithms, namely via equation \eqref{eq:cff} or \eqref{eq:loss}, and \emph{then} find the actions that, given the SCM, would lead to that counterfactual profile?

This program is perfectly pursuable by simply computing \[
a = u^* - u^0,
\]
where $u^*=F^{-1}(x^*)$ and $u^0=F^{-1}(x^0)$ are the residuals with respect to the given SCM of the found counterfactual $x^*$ and the original observation $x^0$, respectively. We refer to this approach as computing the \emph{ex post} actions, and it would indeed save all the effort to translate the model into the latent space, and it would reduce to the much simpler task of computing residuals. The drawback is that the actions found will not, in general, satisfy feasibility constraints, since causality is here considered only in retrospective, and there's nothing preventing the found counterfactual $x^*$ to be unreachable with respect to underlying causal structure.

A toy example will clarify this: suppose you have two variables $A$ and $B$, where, e.g, $B = \alpha A + U_B$, with $\alpha > 0$. Namely, an increase in $A$ causes a linear increase in $B$. Suppose also that, given an observation $x^0$, the probability of finding a valid counterfactual is higher in regions where $A$ is higher but $B$ don't. Then, finding $x^*$ will likely get to a situation where $A$ is higher and $B$ is fixed. In terms of actions, this is only possible when $B$ is intervened upon with a \emph{decrease} in order to compensate for the increase caused by $A$. If, for any reason, there was a feasibility constraint on actions on $B$ such that only increasing interventions are allowed, then the found counterfactual $x^*$ would not correspond to a set of feasible actions.

Therefore, even if it would be in principle possible to compute counterfactual explanations via the standard optimization problem \eqref{eq:loss} in feature space and \emph{then}, given a SCM, compute the corresponding actions to be used as recommendations, this simpler procedure fails to satisfy, in general, feasibility constraints. Of course there are cases in which this simpler approach results in outcomes very similar to the CEILS outcomes: these are the cases in which feasibility constraints ``don't mix with'' underlying causality. Indeed, feasibility problems arise when there are constraints on a variable $B$ having mutable parents ($A$ in the example above). In this case, the changes in $B$ due to the changes in its parents --- that are not ``seen'' by the standard counterfactual optimizer --- may result in an action on $B$ that is no more compatible with the feasibility constraints. We shall discuss more on this in the next section, devoted to experiments.

% ================================================================================================= %
\section{Experiments \label{sec:experiments}}

In this section, we present the experiments conducted on several datasets to validate the CEILS method. We compare our results with a baseline generator of counterfactual explanations using a set of metrics that captures the particularities of our proposal. Next, we describe the datasets used in the experiments and the experiments setup, detailing the definition of the metrics considered. Finally, we discuss the obtained results. 

\subsection{Datasets}\label{sec:datasets}
We use for the experiments a synthetic dataset, two public datasets (German Credit and Sachs) and a proprietary dataset of the financial domain.

\paragraph{Synthetic dataset.}

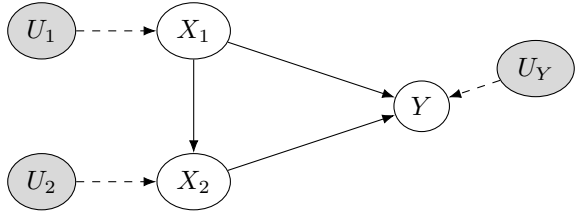
\begin{figure}
\begin{center}
\vspace{3ex}
\begin{tikzpicture}
    % nodes
    %% endogenous
    \node[state] (1) at (-1, 1) {$X_1$};
    \node[state] (2) at (-1, -1) {$X_2$};
    \node[state] (y) at (2, 0) {$Y$};
    %% exogenous
    \node[state, fill=gray!30] (U1) at (-3, 1) {$U_1$};
    \node[state, fill=gray!30] (U2) at (-3, -1) {$U_2$};
    \node[state, fill=gray!30] (Uy) at (3.5, 0.5) {$U_Y$};
% edges
    \path (1) edge (2);
    \path (1) edge (y);
    \path (2) edge (y);
    \path[dashed] (U1) edge (1);
    \path[dashed] (U2) edge (2);
    \path[dashed] (Uy) edge (y);
\end{tikzpicture}
\end{center}
\caption{Causal graph relative to the SE \eqref{eq:SE_synth}.}
\label{fig:synth_graph}
\end{figure}
%\begin{figure*}[htbp]
%    \centering
%    \includegraphics[width=0.5\textwidth]{figs/discussion/graph_syntetic.PNG}
%    \caption{Causal graph of synthetic dataset.}
%    \label{fig:graph_syntetic}
%\end{figure*}

We generate a toy dataset of 100,000 samples with two features ($X_1$ and $X_2$) and a binary outcome ($Y$) with the following Structural Equations: 
\begin{subequations}
\begin{align}
    &X_1 =U_1,\quad     U_1 \sim \mathcal{N}(-1,1)\\
    &X_2 =X_1+U_2,\quad     U_2 \sim \mathcal{N}(5,1)\\
    &Y = \bm{1}_{(3 X_2 - X_1 + U_Y) > t},\quad     U_Y \sim \mathcal{N}(0,1)
\end{align}
\label{eq:SE_synth}
\end{subequations} 
where, in the experiment, $t$ is chosen for simplicity as the median value of $(3 X_2 - X_1 + U_Y)$.

Figure~\ref{fig:synth_graph} depicts the causal graph of the dataset. The rationale behind equations~\eqref{eq:SE_synth} is to have a very simple model that nevertheless allows us to show some crucial aspects of our proposed methodology. In particular, the key feature of equations~\eqref{eq:SE_synth} is to have a non-root node $X_2$ that has a high impact on the target variable $Y$. In this way, we expect to have interesting results when considering $X_2$ a non-actionable feature. 
To this end, we define 2 different experiments: first we set $X_2$ to actionable (Synthetic \#1), then we consider $X_2$ as mutable but non-actionable (Synthetic \#2).

\paragraph{German credit dataset~\citep{german}.}

%\begin{figure*}[htbp]
%    \centering
%    \includegraphics[width=0.75\textwidth]{figs/discussion/graph_german.PNG}
%    \caption{Causal graph of german dataset.}
%    \label{fig:graph_german}
%\end{figure*}

This dataset contains financial information of 1,000 applicants who are classified into applicants with high and low risk of defaulting on their loans. We consider a subset of the features in the same way as \cite{karimi2020algorithmic}. In particular, we use four main features with the causal relations represented in the DAG of figure~\ref{fig:german_graph}. Moreover, we constrain  gender to be immutable and age to increase only.

\paragraph{Sachs dataset \citep{sachs2005causal}.}

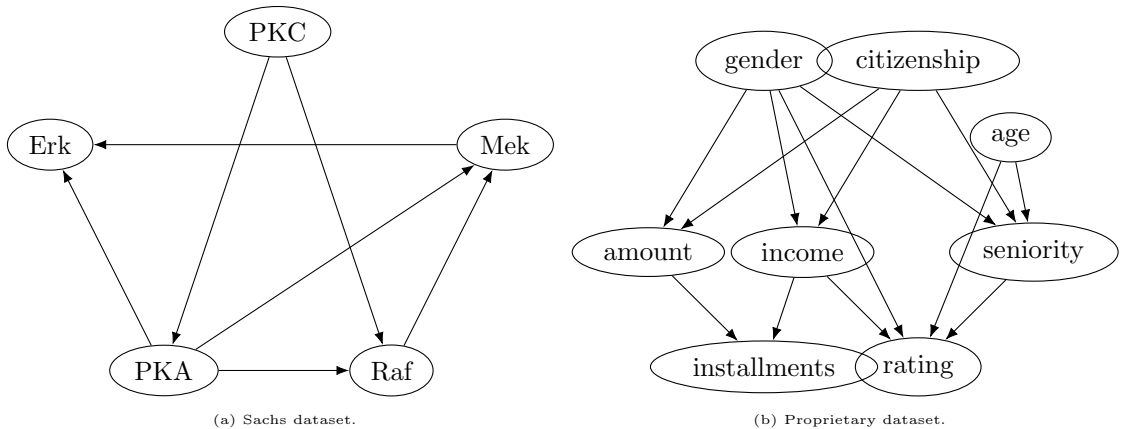
\begin{figure}
     \begin{subfigure}[b]{0.48\textwidth}
          \centering
          \resizebox{\linewidth}{!}{%\begin{figure}
%\caption{Causal graph for the Sachs dataset.}
%\label{fig:sachs_graph}
%\begin{center}
%\vspace{3ex}
\begin{tikzpicture}
    % nodes
    %% endogenous
    \node[state] (pkc) at (0, 2.5) {PKC};
    \node[state] (pka) at (-1.5, -2) {PKA};
    \node[state] (raf) at (1.5, -2) {Raf};
    \node[state] (mek) at (3, 1) {Mek};
    \node[state] (erk) at (-3, 1) {Erk};

    % edges
    \path (pkc) edge (pka);
    \path (pkc) edge (raf);
    \path (pka) edge (mek);
    \path (pka) edge (raf);
    \path (pka) edge (erk);
    \path (mek) edge (erk);
    \path (raf) edge (mek);
\end{tikzpicture}
%\end{center}
%\end{figure}
}  
          \caption{Sachs dataset.}
          \label{fig:sachs_graph}
     \end{subfigure}
     \begin{subfigure}[b]{0.48\textwidth}
          \centering
          \resizebox{\linewidth}{!}{%\begin{figure}
%\caption{Causal graph for the proprietary dataset}
%\label{fig:proprietary_graph}
%\begin{center}
%\vspace{3ex}
\begin{tikzpicture}
    % nodes
    %% endogenous
    \node[state] (c) at (1, 1.5) {citizenship};
    \node[state] (a) at (2.2, 0.5) {age};
    \node[state] (s) at (2.5, -1) {seniority};
    \node[state] (r) at (1, -2.5) {rating};
    \node[state] (in) at (-1, -2.5) {installments};
    \node[state] (am) at (-2.5, -1) {amount};
    \node[state] (i) at (-0.5, -1) {income};
    \node[state] (g) at (-1, 1.5) {gender};
    %\node[state] (y) at (2, -2) {decision};
    
    % edges
    \path (i) edge (in);
    \path (i) edge (r);
    \path (a) edge (s);
    \path (a) edge (r);
    \path (c) edge (i);
    \path (c) edge (s);
    \path (c) edge (am);
    \path (am) edge (in);
    \path (s) edge (r);
    \path (g) edge (am);
    \path (g) edge (i);
    \path (g) edge (s);
    \path (g) edge (r);
\end{tikzpicture}
%\end{center}
%\end{figure}
}  
          \caption{Proprietary dataset.}
          \label{fig:proprietary_graph}
     \end{subfigure}
     \caption{Causal graphs employed in the experiments.}
 \end{figure}

% Description found online
% The data consist in the simultaneous measurements of 11 phosphorylated proteins and phospholipids derived from thousands of individual primary immune system cells, subjected to both general and specific molecular interventions (Sachs et al., 2005).
% Description in Sachs paper
% Each independent sample in this data set consists of quantitative amounts of each of the 11 phosphorylated molecules, simultaneously measured from single cells
% Other descrption
% The flow cytometry data consists of simultaneous measurements of expression levels of 11 biochemical agents in individual cells of the human immune system under 14 different experimental conditions. https://arxiv.org/pdf/1606.07035.pdf
This dataset contains information on protein expression levels in the human immune system. In particular, it consists of 854 observations with 11 independent measurements of phosphorylated molecules derived from immune system cells, subjected to molecular interventions. We base the experiment on the molecules \texttt{PKC}, \texttt{MEK}, \texttt{Raf} and \texttt{PKA} to predict \texttt{Erk}, considering a binary problem with $Y= 1$ when \texttt{Erk} is above the median value, and $Y=0$ otherwise. The variables are related according to the DAG depicted in figure~\ref{fig:sachs_graph} (obtained from \cite{sachs2005causal}). The inhibition and activation of the molecules included in \cite{sachs2005causal} define the constraints that we impose in our experiment. In particular, we consider \texttt{Raf} as non-actionable but mutable, we impose that we can act on \texttt{PKA} only by increasing it and on \texttt{Mek} only by decreasing it. As reported in \cite{McCubrey2007roles}, the molecules \texttt{Raf/Mek/Erk} are associated with growth factors and it could be interesting to control \texttt{Erk}, without intervening directly on it, to prevent cell proliferation and apoptosis.

\paragraph{Proprietary dataset.}

%\input{figs/proprietary_graph}

%\begin{figure*}[htbp]
%    \centering
%    \includegraphics[width=0.75\textwidth]{figs/discussion/ISP_graph.PNG}
%    \caption{Causal graph of Pty dataset.}
%    \label{fig:ISP_graph}
%\end{figure*}

We use a proprietary dataset with 220,304 credit applications~\citep{befair}. This contains 8 features, namely gender, age, citizenship, monthly income, bank seniority, requested amount, number of payments and rating, and the information about the granting/non-granting of the loan. The features are related according to the DAG shown in the figure~\ref{fig:proprietary_graph}. We refer to~\cite{befair} for more details on data and the corresponding causal graph.
We consider gender and citizenship as immutable features and age and bank seniority as features that only can increase in value. Moreover, as in the synthetic case, we run two different experiments, one in which rating is set to actionable (Proprietary \#1), and one in which it is set to mutable but non-actionable (Proprietary \#2).

\subsection{Experiments setup}
For all the experiments, we model the $\{\mathcal{M}_v\}$ as feed-forward neural networks with 2 hidden layers. The classifier $\mathcal{C}$ is also modeled as a feed-forward neural network with 2 hidden layers with ReLU activation functions. We employ the open source library \texttt{TensorFlow} for the implementations~\citep{tensorflow2015-whitepaper}.

\subsubsection{Baseline generator of counterfactual explanations}
We use as baseline generator of counterfactuals the interpretable counterfactual explanations guided by prototypes~\citep{alibi} and in particular, the implementation included in the open source library \texttt{Alibi}~\citep{alibisoft}. More specifically, we implement a counterfactual generator with loss weights: 0.2 for $L_y$ (\texttt{kappa}), 100 for $L_\text{prototype}$ (\texttt{theta}), 0.5 for the $L_1$ proximity term (\texttt{beta}), and 0.5 for the $L_{k-d\ \text tree}$ term (\texttt{gamma}). We employ the $k$-$d$ tree term instead of the autoencoder option. We refer to \cite{alibisoft} for details on these parameters.

%CounterFactualProto(M, shape, use_kdtree=True, theta=100, \# L_proto term, kappa=.2, \# attack loss term L_pred, beta=.5, \# L1 term, gamma=.5, \# L_ae term (autoencoder or kd-tree), max_iterations= 500, learning_rate_init=1e-3, feature_range=(a, b), c_init=1., _steps=10, sess=sess)

We obtain counterfactual explanations first by straight application of the counterfactual generator guided by prototypes (referred to as ``Proto'' in Table~\ref{tab:results}), and then by overlying our CEILS procedure. 

We evaluate these 2 sets of counterfactual explanations by means of a collection of metrics discussed in section~\ref{sec:metrics}. These metrics are of course computed only on valid counterfactuals. Notice, however, that the set of valid counterfactual explanations is, in general, dependent on the methodology. Therefore, we also compute the values of the metrics on the \emph{intersection} of valid counterfactuals explanations obtained from of both methodologies. This is done in order to have a fair comparison, since one of the methods could have, e.g., very good metrics on very few valid counterfactual explanations, namely the ones related to the factual profiles that are the easiest to be explained.

The evaluation is performed on counterfactual explanations generated for 100 random out-of-sample observations for the German credit dataset, 80 for the Sachs dataset, and on 1,000 for the rest of datasets.

\subsubsection{Evaluation metrics}
\label{sec:metrics}

Table~\ref{tab:results} shows our experimental results with respect to the following set of metrics (see e.g. \cite{verma2020counterfactual} and \cite{karimi2021survey} for a review of potential metrics):

\begin{itemize}
    \item \emph{Validity} is the fraction of generated explanations that are valid counterfactual explanations, i.e. such that $\mathcal{C}(x^{\text{cf}}) \neq \mathcal{C}(x)$. Thus, it reflects the effectiveness of a method in generating explanations.

    \item \emph{Proximity}, as discussed in section~\ref{sec:prob_setting}, measures how far is the counterfactual explanation from the original instance. Following~\cite{wachter2017counterfactual} and \cite{dice}, the proximity of continuous features is computed as the mean of feature-wise $L_1$ distance re-scaled by the Median Absolute Deviation from the median (MAD) of each feature. On the other hand, for categorical features we consider a distance of 1 whenever the counterfactual example differs form the original input:
    
    \begin{subequations}
        \begin{align}
            &proximity_{\text{cont}}(x^{\text{cf}},x) = \frac{1}{n\textsubscript{cont}} \sum_{p=1}^{n_\text{cont}} 
            \frac{|{x^{\text{cf}}_p - x_p}|}{MAD_p},
            \\
            &proximity_{\text{cat}}(x^{\text{cf}},x)  = \frac{1}{n_\text{cat}} \sum_{p=1}^{n_\text{cat}} 
            I(x^{\text{cf}}_p \neq x_p);
        \end{align}
        \label{eq:proximity}
    \end{subequations}
    
    where $n_\text{cont}$  and $n_\text{cat}$ are the number of continuous and categorical features respectively, and $MAD_\text{p}$ is the Median Absolute Deviation from the median for the $p$-th continuous variable.

    \item \emph{Sparsity} measures the number of features changes that distinguish the counterfactual explanation from the original instance. In particular, to identify relevant perturbations we consider a threshold as in \cite{dice}: 
    $t = \min(MAD(f), q_{10}(|\w{f}-\text{median}(\w{f})|))$, $\w{f}=[f_i: f_i \neq \text{median}(f)]$\\
    
    \begin{equation}
        sparsity(x^{\text{cf}}, x) = \sum_{p=1}^d \bm{1}_{\left\{\rvert x^{\text{cf}}_p - x_p\rvert > t_p\right\}}.
    \end{equation}
    
    Analogously, we compute sparsity also in terms of actions (see below).

    \item \emph{Distance}, related to the proximity, measures the $L_1$ distance between counterfactual and factual observations:
    \begin{equation}
        distance(x^{\text{cf}}, x) = \lVert x^{\text{cf}} - x \rVert_1.
        \label{eq:distance}
    \end{equation}
\end{itemize}

All the above metrics refer directly to the counterfactual explanations, thus we refer to them as metrics on feature space. Instead, the following metrics are focused on the evaluation of actions, thus on latent space quantities. Notice that, when considering the baseline method Proto (or, in general, non-causal methods) there is no latent space to be considered, or, equivalently as discussed above, actions are all hard interventions, i.e. shifts in feature space. However, as argued in section~\ref{sec:action_feature_space}, we could alternatively think of generating counterfactuals with a non-causal method and then compute anyway the corresponding \emph{ex post} action via the SCM. In the section of table~\ref{tab:results} devoted to latent space metrics, for rows corresponding to the baseline method we have indeed reported metrics computed with this \emph{ex post} rationale, thus computing the residuals of the generated counterfactual explanations with respect to the SCM.       

\begin{itemize}

    \item \emph{Cost}, as discussed in section~\ref{sec:recourse}, is the magnitude of the action needed to reach a counterfactual point. Specifically, we compute the cost as the $L_1$ norm of the action: $\lVert a\rVert_1$. In terms of the feature space, considering the SCM we have: 
    \begin{equation}
        cost(x^{\text{cf}}, x) = \lVert F^{-1}(x^{\text{cf}}) - F^{-1}(x)\rVert_1.
        \label{eq:cost_causal}
    \end{equation}
    Notice that this is in fact equivalent to the \emph{distance} metric, but on the latent space.

    \item \emph{Feasibility}, as discussed in section~\ref{sec:recourse}, pertains to the fact that suggested actions are realistic, i.e. actually doable. Therefore, it includes all the requirements upon actions. For example, when a feature is non-actionable any non-null action on that feature is unfeasible. In the same way, when a feature can only increase (e.g. age), any negative action on that feature would be unfeasible. We measure feasibility as the percentage of explanations whose actions are all feasible. Thus, we establish the following formulation:
    
    \begin{subequations}
    \begin{align}
        & F^{-1}(x^\text{cf})_v = F^{-1}(x)_v \iff a_v = 0,\quad \forall v\ \text{non-actionable};\\
        & F^{-1}(x^\text{cf})_v \geq F^{-1}(x)_v\iff a_v \geq 0,\quad \forall v\ \text{increasing only};\\
        & F^{-1}(x^\text{cf})_v \leq F^{-1}(x)_v\iff a_v \leq 0,\quad \forall v\ \text{decreasing only}.
    \end{align}
    \end{subequations}
    
    \item \emph{Causal plausibility} is inspired by \cite{mahajan2019preserving}, who propose to add this term to the loss in problem \eqref{eq:loss} in order to keep the generated explanations ``as close as possible'' to the underlying causal structure. The intuition is to measure, for each feature, the distance between the found counterfactual observation and the value that the feature should have if perfectly obeying to the SCM, i.e. with null residuals. The idea is to compare $x^\text{cf}_v$ with $f_v(\pa(x^\text{cf}_v))$. To compute this, one has to build the vector 
    \begin{equation*}
        w_v = \left\{
            \begin{aligned} 
                & \mathcal{M}_v(\pa(x^\text{cf}_v)), \quad v\ \text{non root}\\
                & x^\text{cf}_v,\quad v\ \text{root}
            \end{aligned}
            \right.
    \end{equation*} 
    then: \begin{equation}
        causal\ plausibility(x^{\text{cf}}, x) = \lVert x^{\text{cf}} - w \rVert_1. 
    \end{equation}
    Notice that this is equivalent to computing the $L_1$ norm of the (non-root) residuals of $x^{\text{cf}}$ with respect to the SCM.
    In other words, this metric measures the distance of the found counterfactual from the profile that satisfies the SCM with zero residuals (except for root nodes).
\end{itemize}

Moreover, we compute one more value designed not to compare a counterfactual explanation $x^{\text{cf}}$ with its factual counterpart $x$ --- as the ones introduced above --- but rather to directly compare two methods for counterfactual generation. In particular, we are interested in comparing our proposed methodology with the baseline methodology to understand the net impact of our approach on the underlying counterfactual generator engine. 
To this end, we compute  \mbox{$\lVert (x^{\text{cf, base}} - x) - a^\text{CEILS}\rVert_1$} --- where $(x^{\text{cf, base}} - x)$ is the recommended action by the baseline generator and $a^\text{CEILS}$ is the action proposed by our methodology --- to measure whether the two methods are recommending actions in the same direction. Obviously, this metric can be computed only on valid counterfactual explanations common to both methods.

\subsection{Results}\label{sec:results}

Table \ref{tab:results} summarizes the results obtained in the experiments for all the datasets and metrics. As mentioned, we first compute metrics for the explanations obtained with the two methodologies (Proto and Proto + CEILS) and then for the valid explanations common to both methods (grey rows in the Table~\ref{tab:results}).

Except for validity and feasibility, for each metric we report the median value and the deviation from the median computed over valid counterfactual explanations found by the corresponding method. As mentioned, we include the same computation over valid counterfactuals common to both methods (grey rows in the table).

In what follows, we first describe the results obtained for each dataset, pointing out the main findings, then we summarize them in section~\ref{sec:discussion}.

\begin{table}
\centering 
\caption{\textbf{Results of the experiments.} The table reports the value of nine metrics for six experiments using four datasets (i.e. Synthetic dataset, German credit dataset, Sachs dataset, and Proprietary dataset) to compare the counterfactual explanations generated via prototypes by \cite{alibi} (Proto) with our proposal on top of the same counterfactual generator (Proto + CEILS). Validity and feasibility are reported as percentages, while for all the other metrics we report the median values and corresponding deviations from the median. Values are computed over standardized feature values, except proximity and sparsity, which involve MAD computation of features.
White rows contain values computed on all valid explanations generated by the corresponding method, gray rows display the metrics computed on the intersection of valid explanations of the 2 methods. For the intersection evaluation, in the validity column we report the number of common valid explanations.}

\label{tab:results} 

\ra{1.4}
\resizebox{\textwidth}{!}{
    \begin{tabular}{lccccccccccccc}
    
    \toprule
    %first row -titles
    &&&&& \multicolumn{4}{c}{\textbf{feature space}}&&\multicolumn{4}{c}{\textbf{latent space}}\\
    \cmidrule{6-9}\cmidrule{11-14}
    & \phantom{a}
    & \textbf{method}
    & \textbf{validity}
    & \phantom{a}
    & \makecell{\textbf{continuous}\\ \textbf{proximity}}
    & \makecell{\textbf{categorical}\\ \textbf{proximity}}
    & \textbf{sparsity} 
    & \textbf{distance}
    & \phantom{a} 
    & \makecell{\textbf{sparsity} \\ \textbf{on actions}}
    & \textbf{cost}
    & \textbf{feasibility} 
    & \makecell{\textbf{causal} \\ \textbf{plausibility}} 
    \\ 
    
    \cmidrule{3-14}
    
    %Synthetic data #1
    \multirow{4}{*}{\makecell[l]{Synthetic \#1 \\ ($X_2$ actionable)}} 
        && Proto               
        & 100\% &
        & 0.64 $\pm$ 0.49                                     
        & -        
        & 1.34 $\pm$ 0.61       
        & 0.65 $\pm$ 0.44           
        &
        & 1.34 $\pm$ 0.61 
        & 0.65 $\pm$ 0.48 
        & 100\%
        & 0.43 $\pm$ 0.22
        \\
    
        && Proto + CEILS               
        & 99\% &        
        & 0.78 $\pm$ 0.61     
        & -        
        & 1.48 $\pm$ 0.63
        & 0.88 $\pm$ 0.61           
        &
        & 1.47 $\pm$ 0.64
        & 0.73 $\pm$ 0.46
        & 100\%                                                         
        & 0.33 $\pm$ 0.18
        \\ 
        \\[-1ex]
    
        \rowcol \cellcolor{white}&\cellcolor{white}& Proto 
        & \#999  &    
        & 0.64 $\pm$ 0.49       
        & -     
        & 1.34 $\pm$ 0.61      
        & 0.64 $\pm$ 0.44           
        &
        & 1.34 $\pm$ 0.61               
        & 0.64 $\pm$ 0.48 
        & 100\%
        & 0.43 $\pm$ 0.22
        \\
    
        \rowcol \cellcolor{white}&\cellcolor{white}&Proto + CEILS   
        & \#999 &
        & 0.78 $\pm$ 0.61                
        & -       
        & 1.48 $\pm$ 0.63        
        & 0.88 $\pm$ 0.61         
        &
        & 1.47 $\pm$ 0.64     
        & 0.73 $\pm$ 0.46         
        & 100\%      
        & 0.33 $\pm$ 0.18
        
        \\
        
    \cmidrule{3-14}
    
    %Synthetic data #2
    \multirow{4}{*}{\makecell[l]{Synthetic \#2 \\ ($X_2$ non-actionable)}} 
        && Proto               
        & 63.1\% &
        & 1.4 $\pm$ 0.83                                     
        & -        
        & 1.0 $\pm$ 0.0       
        & 1.73 $\pm$ 0.88           
        &
        & 2.0 $\pm$ 0.0
        & 2.93 $\pm$ 1.49                             
        & 0\%                                                           
        & 0.99 $\pm$ 0.43
        \\
    
        && Proto + CEILS               
        & 99.5\%  &   
        & 1.98 $\pm$ 1.4     
        & -        
        & 1.96 $\pm$ 0.23
        & 2.31 $\pm$ 1.29           
        &
        & 0.99 $\pm$ 0.08 
        & 1.34 $\pm$ 0.76                              
        & 100\%                                                         
        & 0.48 $\pm$ 0.28
        \\
        \\[-1ex]

        \rowcol \cellcolor{white}&\cellcolor{white}& Proto   
        & \#631   &   
        & 1.4 $\pm$ 0.83       
        & -     
        & 1.0 $\pm$ 0.0      
        & 1.73 $\pm$ 0.88           
        &
        & 2.0 $\pm$ 0.0                
        & 2.93 $\pm$ 1.49   
        & 0\%
        & 0.99 $\pm$ 0.43
        \\
    
        \rowcol \cellcolor{white}&\cellcolor{white}&Proto + CEILS
        & \#631&
        & 1.14 $\pm$ 0.68                
        & -       
        & 1.94 $\pm$ 0.28        
        & 1.41 $\pm$ 0.69         
        &
        & 0.99 $\pm$ 0.1     
        & 0.83 $\pm$ 0.4         
        & 100\%        
        & 0.39 $\pm$ 0.23
        
        \\
    \cmidrule{3-14}
        
    % German dataset
    \multirow{4}{*}{German credit}
        && Proto               
        & 70\%   &    
        & 1.88 $\pm$ 0.57             
        & 0.0 $\pm$ 0.0                  
        & 2.44 $\pm$ 0.67        
        & 2.94 $\pm$ 0.53        
        &
        & 2.5 $\pm$ 0.63              
        & 2.89 $\pm$ 0.52
        & 100\% 
        & 1.6 $\pm$ 0.13          
        \\
        
        && Proto + CEILS               
        & 77\% &
        & 1.97 $\pm$ 0.64 
        & 0.0 $\pm$ 0.0   
        & 2.4 $\pm$ 0.69    
        & 2.99 $\pm$ 0.54        
        &
        & 2.37 $\pm$ 0.7 
        & 2.89 $\pm$ 0.54           
        & 100\% 
        & 1.6 $\pm$ 0.18          
        \\ 
        \\[-1ex]
    
        \rowcol \cellcolor{white}&\cellcolor{white}& Proto              
        & \#67 &       
        & 1.87 $\pm$ 0.58                                         
        & 0.0 $\pm$ 0.0      
        & 2.43  $\pm$ 0.68      
        & 2.91 $\pm$ 0.51      
        &
        & 2.49 $\pm$ 0.64                
        & 2.89 $\pm$ 0.52  
        & 100\%                                                    
        & 1.6 $\pm$ 0.12          
        \\

        \rowcol \cellcolor{white}&\cellcolor{white}& Proto + CEILS    
        & \#67  &   
        & 1.86 $\pm$ 0.57 
        & 0.0 $\pm$ 0.0  
        & 2.45 $\pm$ 0.68    
        & 2.91 $\pm$ 0.51       
        &
        & 2.4 $\pm$ 0.7  
        & 2.87 $\pm$ 0.55
        & 100\% 
        & 1.59 $\pm$ 0.16          
        \\

        \cmidrule{3-14}
    %Sachs dataset
    \multirow{4}{*}{\makecell[l]{Sachs \\}}
        && Proto               
        & 37.5\% &
        & 2.08 $\pm$ 0.75
        & -
        & 2.1 $\pm$ 0.66
        & 3.07 $\pm$ 0.95
        &
        & 3.63 $\pm$ 0.81
        & 4.13 $\pm$ 0.96
        & 0\%
        & 2.36 $\pm$ 0.81
        \\
    
        && Proto + CEILS               
        & 22.5\%   &
        & 2.35 $\pm$ 0.86
        & -
        & 3.56 $\pm$ 0.98
        & 3.47 $\pm$ 0.68
        &
        & 1.91  $\pm$  0.58
        & 2.5 $\pm$  0.69
        & 100\%
        & 2.17 $\pm$   0.5
        \\ 
        \\[-1ex]
        
        \rowcol \cellcolor{white}&\cellcolor{white}& Proto                            
        & \#18 &
        & 1.83 $\pm$ 0.73
        & -
        & 1.83 $\pm$ 0.62
        & 2.64  $\pm$  0.7
        &
        & 3.44 $\pm$ 0.98
        & 3.93  $\pm$  0.84
        & 0\%
        & 2.77 $\pm$  0.87
        \\
    
        \rowcol \cellcolor{white}&\cellcolor{white}&Proto + CEILS 
        & \#18 &
        & 2.35 $\pm$ 0.86
        & -
        & 3.56 $\pm$ 0.98
        & 3.47  $\pm$  0.68
        &
        & 1.89  $\pm$  0.68
        &  2.5  $\pm$  0.69
        & 100\%
        & 2.17  $\pm$ 0.5

        \\

    \cmidrule{3-14}

    %ISP dataset rating actionable
    \multirow{4}{*}{\makecell[l]{Proprietary \#1 \\ (rating actionable)}}
        && Proto               
        & 100\% &
        & 1.49 $\pm$ 9.31
        & 0.32 $\pm$  0.06
        & 1.46 $\pm$ 0.69
        & 1.26 $\pm$ 0.65
        &
        & 1.68 $\pm$ 0.98
        & 1.36 $\pm$ 0.69
        & 98\%
        & 3.32 $\pm$ 0.7
         \\
    
        && Proto + CEILS               
        & 100\% &  
        & 4.12 $\pm$ 17.49
        & 0.32 $\pm$ 0.05
        & 1.65 $\pm$ 1.05
        & 1.39 $\pm$ 0.69
        &
        & 1.43 $\pm$ 0.75
        & 1.26 $\pm$ 0.64
        & 100\%
        & 3.33 $\pm$ 0.71
        \\ 
        \\[-1ex]
    
        \rowcol \cellcolor{white}&\cellcolor{white}& Proto               
        & \#1000 &
        & 1.49 $\pm$ 9.31
        & 0.32 $\pm$  0.06
        & 1.46 $\pm$ 0.69
        & 1.26 $\pm$ 0.65
        &
        & 1.68 $\pm$ 0.98
        & 1.36 $\pm$ 0.69
        & 98\%
        & 3.32 $\pm$ 0.7
        \\
    
        \rowcol \cellcolor{white}&\cellcolor{white}& Proto + CEILS 
        & \#1000 &
        & 4.12 $\pm$ 17.49
        & 0.32 $\pm$ 0.05
        & 1.65 $\pm$ 1.05
        & 1.39 $\pm$ 0.69
        &
        & 1.43 $\pm$ 0.75
        & 1.26 $\pm$ 0.64
        & 100\%
        & 3.33 $\pm$ 0.71

        \\ 

        \cmidrule{3-14}

    %ISP dataset rating not actionable
    \multirow{4}{*}{\makecell[l]{Proprietary \#2 \\ (rating non-actionable)}}
        && Proto               
        & 21.8\% &
        & 289.57 $\pm$ 830.79
        & 0.0 $\pm$ 0.0
        & 2.86 $\pm$ 0.95
        & 2.16 $\pm$ 1.1
        &
        & 3.49 $\pm$ 1.08
        & 2.51 $\pm$ 1.24
        & 6\%
        & 3.68 $\pm$ 1.33
        \\
    
        && Proto + CEILS               
        & 81.6\%   &
        & 474.15 $\pm$ 879.79
        & 0.12 $\pm$ 0.16
        & 3.18 $\pm$ 0.96
        & 8.63 $\pm$ 5.33
        &
        & 2.71 $\pm$ 0.76
        & 8.65 $\pm$ 5.74
        & 100\%
        & 10.18 $\pm$ 5.29
        \\ 
        \\[-1ex]
        
        \rowcol \cellcolor{white}&\cellcolor{white}& Proto                            
        & \#218 &
        & 289.57 $\pm$ 830.79
        & 0.0 $\pm$ 0.0
        & 2.86 $\pm$ 0.95
        & 2.16 $\pm$ 1.1
        &
        & 3.49 $\pm$ 1.08
        & 2.51 $\pm$ 1.24
        & 6\%
        & 3.68 $\pm$ 1.33
        \\
    
        \rowcol \cellcolor{white}&\cellcolor{white}&Proto + CEILS 
        & \#218 &
        & 43.23 $\pm$ 109.46
        & 0.09 $\pm$ 0.15
        & 2.83 $\pm$ 1.17
        & 1.72 $\pm$ 0.87
        &
        & 2.28 $\pm$ 1.04
        & 1.35 $\pm$ 0.81
        & 100\%
        & 3.3 $\pm$ 1.2

        \\

    \bottomrule
    \end{tabular}
}
\end{table}

\paragraph{Synthetic dataset.} 
For this dataset we run 2 different experiments, which differ only for the actionability of $X_2$. As discussed in section~\ref{sec:datasets}, $X_2$ is the feature with highest impact on the target $Y$, and it is causally dependent on $X_1$. Therefore, it is interesting to assess our methodology when $X_2$ is mutable but non-actionable: CEILS should be able to learn how to employ $X_1$ in order to impact $X_2$ and thus $Y$; on the other hand, the baseline Proto cannot, this either treats $X_2$ as actionable, but in this way this suggests the unfeasible action of changing $X_2$, or as non-actionable. However, in this case, the baseline Proto method will not be able to move $X_2$ in any way, hardly resulting in efficient counterfactual generations. 

Even if we inspect what actions would have been recommended for the baseline model considering \emph{ex post} actions (see section~\ref{sec:action_feature_space}), both strategies ($X_2$ actionable or non-actionable) result in non-feasible suggestions with respect to $X_2$: for the actionable case $X_2$ is changed by the baseline method, but very likely in a way that is not compensated by the change in the parent $X_1$, resulting in unfeasible net action on $X_2$; for the non-actionable case $X_2$ is kept fixed by design in the baseline method, but a non-null action in $X_2$ is unavoidable to keep $X_2$ fixed while changing $X_1$, which is unfeasible. 

These considerations are confirmed by the experimental results in table~\ref{tab:results} and in the examples shown in table~\ref{tab:synth_examples}.

\begin{table}
    \centering
    \caption{\textbf{Examples of explanations}. Two examples of instances of the Synthetic~\#2 experiment explained by counterfactual profiles generated via the prototype method of~\cite{alibi} and via CEILS. The $\Delta$ stands for the difference between counterfactual and factual instances. The action for the baseline method is the \emph{ex post} action.}\label{tab:synth_examples}
    \ra{1.2}
    \resizebox{.7\textwidth}{!}{ 
    \begin{tabular}{lccccccccccc}
      \toprule
      &&& \phantom{a} & \multicolumn{2}{c}{\bf counterfactuals} & \phantom{a} & \multicolumn{2}{c}{\bf $\Delta$} & \phantom{a} & \multicolumn{2}{c}{\bf Action}\\
      \cmidrule{5-6}\cmidrule{8-9}\cmidrule{11-12}
      &\bfseries Variable & \bfseries Factual && \bfseries Proto & \bfseries Proto + CEILS && \bfseries Proto & \bfseries Proto + CEILS && \bfseries Proto & \bfseries Proto + CEILS\\
      \midrule

      \multirow{3}{*}{\textit{Example 1}} & $Y$ & 0 && 1  & 1 \\
      & $X_1$ & -1.098 && -1.364 & -0.995 && -0.266     & 0.103  && -0.260 & 0.103\\

      & $X_2$ & 3.896 && 3.896 & 4.006 && 0 & 0.110  && 0.242 &  0\\
      
      \midrule
      
      \multirow{3}{*}{\textit{Example 2}} & $Y$ & 1 && 0 & 0\\        
      &$X_1$ & -0.886 && -0.726 & -0.973 && 0.159 & -0.087  && 0.162 & -0.087\\       

      &$X_2$ & 4.080  && 4.080 & 3.999 && 0 & -0.081 && -0.168 & 0\\

      \bottomrule
    \end{tabular}
    }
\end{table}

Indeed the overall results relative to Synthetic \#1 and \#2 in table~\ref{tab:results} show that:
\begin{itemize}
    \item in terms of feature space metrics the Proto method performs slightly better, and this is in line with the fact that CEILS focuses on \emph{nearest actions} rather than nearest explanations;
    \item if we compare the CEILS cost with the effort done by the baseline, namely $\lVert x^\text{cf} - x\rVert_1$ (i.e. the distance metric), then the gain of using CEILS becomes evident in both runs.
    
    \item even if we consider the \emph{ex post} action for the baseline Proto method, we can see that for Synthetic~\#2 ($X_2$ non-actionable) there is a huge gain in cost, and this is due to the fact that the Proto method pushes $X_1$ in the ``wrong'' direction, since it employs causality only \emph{after} computing explanations; 
    
    \item analogously, feasibility metric confirms our expectations: in the Synthetic~\#2 experiment the baseline suggest only unfeasible actions.
    
    \item causal plausibility, much higher for the baseline in Synthetic \#2,  confirms once more this evidence.
    
    \item in Synthetic~\#1, instead, both cost, feasibility and causal proximity are comparable: in this case $X_2$ is actionable, therefore there is no feasibility issue for neither of the two methods.
\end{itemize}

Table~\ref{tab:synth_examples} displays 2 examples of counterfactual explanations for the case in which $X_2$ is retained non-actionable.
If we focus on the first example where \mbox{$Y=0 \rightarrow Y=1$}, the Prototype method tends to decrease $X_1$ (-0.266) because the model $\mathcal{C}$ has learnt the negative dependence of $Y$ in $X_1$, and also it can not act on $X_2$. However, CEILS ``knows'' that by decreasing $X_1$ there is a linear decrease in $X_2$, which has a stronger impact on the outcome $Y$. Indeed, in the action column, we see that CEILS effectively suggests to increase $X_1$ (0.103) in order to increase the quantity $3 X_2 - X_1$, and it does so with less overall effort required from the end user.

Moreover, notice that, in line with expectations, if we compute the SCM \emph{ex post} action for the baseline, then we have non-feasibility for both examples, since there is a non-null action on $X_2$. Also in this case, we see that actions on $X_1$ are done in the ``wrong'' direction.

Similar arguments can be made for the example 2 included in the table~\ref{tab:synth_examples} where the  counterfactual methods try to modify the target as \mbox{$Y=1 \rightarrow Y=0$}.

\paragraph{German credit dataset.}
As shown in table~\ref{tab:results}, the results do not present any significant difference among the 2 methods. This is not surprising, since the only feasibility constraints are on root nodes (gender kept immutable and age not decreasing).
Also in terms of effort there is no apparent discrepancy: the distance obtained with the Proto method and the cost of CEILS are almost identical, i.e. there is no gain given by the causal flow. Also in terms of direction, the two methods seem comparable: \mbox{$\lVert (x^{\text{cf, base}} - x) - a^\text{CEILS}\rVert_1$} has a median value of 0.16.
This may be due to the fact that the causal impact of age on amount and (then) duration is not strong enough to play a significant role. Similarly, \emph{ex post} actions have the same overall cost with respect to CEILS actions.

This experiment shows that the advantage of employing causality on top of standard approaches is not always appreciable, and is highly dependent on the underlying causality structure and on the constraints set over the variables.

%In German dataset there are not significant difference in metrics because the actionable variables are root nodes. The proximity of categorical is 0 because the sex is immutable. Action and distance do not differ from magnitude and direction.

\paragraph{Proprietary dataset.}
For the proprietary dataset, the behavior of the two methods in not dissimilar from the synthetic case, however this experiment involves much more complex causal relationships, and presents a real-world scenario of credit lending. Here, the role of the feature $X_2$ is played by the feature rating, i.e. a feature extremely important to determine the final outcome, that cannot be controlled directly by the end users, and usually is a complex function of other variables.

Similarly to the synthetic dataset, we run two experiments: Proprietary~\#1, where we consider rating actionable, and Proprietary~\#2, where rating is mutable but non-actionable\footnote{Recall that for methods not accounting for causality, non-actionable is effectively equivalent to immutable, since each feature can change only via direct interventions (see section~\ref{sec:action_feature_space}).}.
The results are in line with the discussion made for the synthetic case, thus we here focus only on some interesting insights peculiar to this case:
\begin{itemize}
    \item in Proprietary~\#2 we see that the baseline Proto method is much less efficient than CEILS in providing valid counterfactuals: this is due to the high importance of rating in determining the target variable and to the fact that that the Prototype method cannot change rating indirectly, as CEILS does; 
    \item this also explains the odd discrepancy in terms of costs 8.65 for CEILS vs 2.51 for the baseline: this is indeed an artefact of the small number of valid counterfactuals over which this metric is computed for the baseline method, CEILS has higher cost simply because is finding counterfactuals also for ``harder'' cases. If we compute the metrics on the common valid explanations (grey rows in the table), the situation is reversed.
    \item In Proprietary~\#2 we have $\lVert (x^\text{cf, base} -x) - a^\text{CEILS}\rVert_1 = 1.87$, confirming the fact that the two methods suggest recommendations in very different directions.
\end{itemize}

Table~\ref{tab:propietary_examples} shows an example of explanations generated by both methods in the Proprietary~\#2 setting. As expected gender and citizenship remain fixed, while age and seniority have equal or higher values with respect to the original instance. The baseline method produces a counterfactual explanation with values far away from the factual profile (i.e. increases the income to 3643.3K and almost doubles the requested amount  while decreasing the number of installments). On the other hand, CEILS only suggests to increase the bank seniority and the requested amount\footnote{Both methods apparently provide the counter-intuitive suggestion of increasing the requested amount: this is due to the fact that the baseline method of \cite{alibi} searches for explanations as close as possible to the training data distribution. In other words, a too small requested amount would not be plausible with respect to the other suggested features.}. Indeed, increasing the bank seniority results in better rating, which is enough to reach the loan approval. Evidently, an increase in seniority is impossible without a corresponding increase in age: actually, we have treated bank seniority as an actionable feature, but it would have been more appropriate to consider it as mutable only as a consequence of age changes, since seniority cannot be controlled independently of age, or to consider an additional common confounder. Nevertheless, we have decided to keep seniority actionable to focus our discussion on rating and not to limit too much the baseline method (for which it would have been impossible to change seniority as well as rating).
Moreover, notice that considering the \emph{ex post} action for the prototype method results in having a net \emph{increase} in rating (i.e worsening). This confirms what discussed in section~\ref{sec:action_feature_space}, i.e. to keep the rating fixed the non-causal method needs to intervene with a negative action to compensate the change due to the impact of the suggested changes in other features.

%Figure~\ref{fig:rating_dist} shows the distribution of interventions on rating for 3 scenarios: the baseline Proto method for rating considered either actionable or non-actionable, CEILS with rating non-actionable. It is clear that \emph{ex post} actions for the baseline methods are completely incompatible with the non-actionability of rating, while CEILS easily satisfies this constraint. If we let the rating free, then the baseline method prescribes even larger interventions on ratings, i.e. the \emph{ex post} casual flow is not sufficient to account for the rating changes in the explanations. This shows that for the baseline method there is no way of producing counterfactual profiles that do not imply, \emph{ex post}, a direct intervention on rating, neither if we fix rating nor if we let it free of varying.   

\begin{figure}[htbp]
    \centering
    \includegraphics[width=0.8\textwidth]{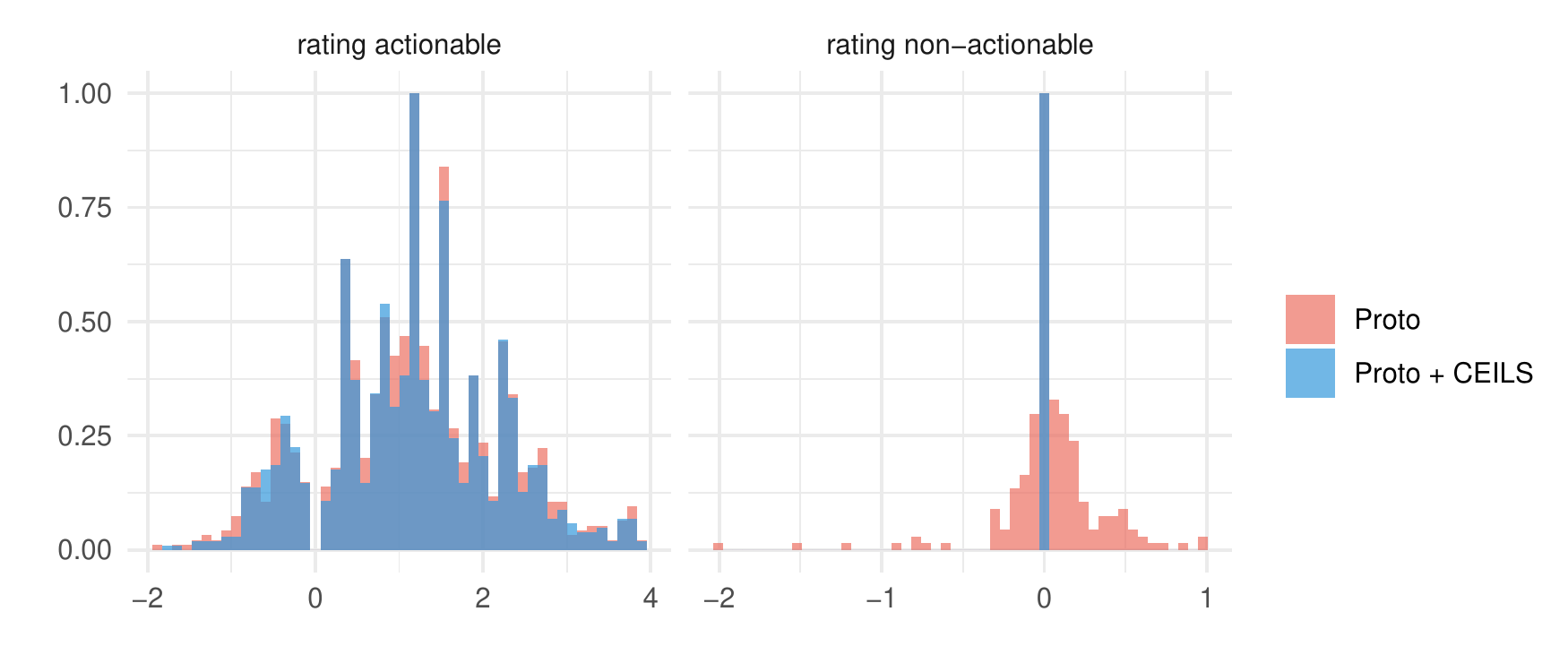}
    \caption{The distribution of actions on rating for 2 scenarios: rating considered either actionable or non-actionable (which, for the baseline method, is equivalent to immutable). For the prototype method the \emph{ex post} actions are reported.  Values are rescaled to a maximum of 1 for sake of readability.}
    \label{fig:rating_dist2}
\end{figure}

Figure~\ref{fig:rating_dist2} shows the distribution of interventions on rating for 2 scenarios: with rating considered either actionable or non-actionable. First, we can notice that the baseline method has non-null \emph{ex post} actions in both cases, thus meaning that non-causal method cannot in any way account for features that are not directly intervened upon but could vary in response to changes in other variables, while CEILS has obviously null actions on rating in the actionable (but mutable) case. Secondly, we see that when rating has no constraints (the actionable case) then the two distributions are very similar: this underlines once again that our approach --- and in general considering causality in counterfactual explanations generator --- gives results in line with its baseline method when there are no feasibility constraints on important features.

\begin{table}
    \centering
    \caption{\textbf{Examples of explanations}. Example of an instance of the Proprietary~\#2 dataset experiment explained by counterfactual profiles generated via the prototype method of~\cite{alibi} and via Proto + CEILS. The $\Delta$ stands for the difference between counterfactual and factual instances. The action for the baseline method is the \emph{ex post} action.}\label{tab:propietary_examples}
    \ra{1.2}
    \resizebox{.7\textwidth}{!}{ 
    \begin{tabular}{lcccccccccc}
      \toprule
      && \phantom{a} & \multicolumn{2}{c}{\bf counterfactuals} & \phantom{a} & \multicolumn{2}{c}{\bf $\Delta$} & \phantom{a} & \multicolumn{2}{c}{\bf Action}\\
      \cmidrule{4-5}\cmidrule{7-8}\cmidrule{10-11}
      \bfseries Variable & \bfseries Factual && \bfseries Proto & \bfseries Proto CEILS && \bfseries Proto & \bfseries Proto + CEILS && \bfseries Proto & \bfseries Proto + CEILS\\
      \midrule

      decision & reject && approve  & approve  \\
      gender & 1 && 1 & 1 && 0 & 0 && 0  & 0\\
      age & 30 && 30 & 30 && 0 & 0 && 0  & 0\\
      citizenship & 1 && 1 & 1 && 0 & 0  && 0& 0\\
      income & 56.7K && 3700K & 56.7K && 3643.3K & 0  && 3643.3K & 0\\
      seniority & 0 && 8.1 & 5.4 && 8.1  & 5.4  && 8.1 & 5.4\\
      amount & 210K && 409K & 320K && 199K  & 110K  && 199K &  110K\\
      installments & 48 && 33.4 & 53.1 && -14.6  & 5.1  && -22.9 & 0\\
      rating & 10 && 10 & 9.625 && 0 & -0.375  && 0.5 &  0\\
      
      %\midrule
    
      %\multirow{3}{*}{\textit{Example 2}} & $Y$ & 1 && 0 & 0\\        
      \bottomrule
    \end{tabular}
    }
\end{table}

\paragraph{Sachs dataset.}
80 random out-of-sample instances were used to generate counterfactual explanations. These result in 30 valid counterfactuals (37.5\%) for the baseline method and 18 (22.5\%) for Proto + CEILS. This can be explained by the fact that Proto + CEILS is satisfying the feasibility constraints (i.e. no action on \texttt{Raf} and monotonic conditions on \texttt{PKA} and \texttt{Mek}) and the compliance to the SCM, while Proto alone is ignoring them, when considering \emph{ex post} actions. In other words, the constraints on actions and the compliance to the SCM make, in this case, CEILS less effective in terms of valid feature space explanations. 

Indeed, the difficulty in providing feasibile counterfactuals can be seen in the intersection part. Proximity, sparsity and distance (metrics in feature space) are higher in Proto + CEILS, while sparsity on action, cost and causal plausibility (metrics in latent space) are all worse for the baseline. Note that Proto + CEILS effort (i.e. cost) (2.50) is slightly lower with respect both to Proto distance (2.64) and also to Proto \emph{ex post} cost, meaning that there is a gain in the effort to reach the suggested explanation.

Inspection of table~\ref{tab:sachs_examples} clearly reveals that the change in feature space ($\Delta$) in \texttt{PKC} and \texttt{PKA} is comparable in the two methods, but CEILS, forced to satisfy the non-actionability in \texttt{Raf} and the non-increase in \texttt{Mek}, has an effective \emph{decrease} in \texttt{Raf} and \texttt{Mek} that the baseline has not. In other words, in this experiment we see that feasibility may also be a friction with respect to providing valid counterfactuals, as it could be expected.
Moreover, looking at \texttt{Mek} (31.2) and \texttt{Raf} (18.0) values for Proto \emph{ex post} actions, we have once more the confirmation that non-causal methods are not able to account for feasible recommendations with respect to a given SCM.

Notice that the baseline method does have the feasible constraints as well, but they are interpreted as hard interventions --- as discussed in \ref{sec:recourse}. Thus, as we see by the example in table~\ref{tab:sachs_examples}, Proto keeps \texttt{Raf} and \texttt{Mek} fixed, but it can change the other features independently of these constraints, while CEILS can not, since any change in the variables has impacts on others as by the SCM. In this sense, CEILS is more constrained than its baseline method.

%Inspection of table~\ref{tab:sachs_examples} clearly reveals that optimising over the latent space (i.e. Proto + CEILS) provides sparser actions while the causal influence leads to a $\Delta$ less sparse in the feature space. Reverse reasoning can be done for the Proto algorithm. Here the quality suggestion, between the two methods, is to inhibit \texttt{PKC} and activate \texttt{PKA} but the difference lays on the quantitative suggestion because CEILS takes into account the causal influence among the variables after the intervention and, as a result, $\Delta$ Proto + CEILS is non sparse but less effort in Action is required compared to $\Delta$ Proto. 
\begin{table}
    \centering
    \caption{\textbf{Examples of explanations}. Example of an instance of the Sachs dataset experiment explained by counterfactual profiles generated via the prototype method of~\cite{alibi} and via Proto + CEILS. The $\Delta$ stands for the difference between counterfactual and factual profiles. The action for the baseline method is the \emph{ex post} action.}\label{tab:sachs_examples}
    \ra{1.2}    \resizebox{.7\textwidth}{!}{ 
    \begin{tabular}{lccccccccccc}
      \toprule
      &&& \phantom{a} & \multicolumn{2}{c}{\bf counterfactuals} & \phantom{a} & \multicolumn{2}{c}{\bf $\Delta$} & \phantom{a} & \multicolumn{2}{c}{\bf Action}\\
      \cmidrule{5-6}\cmidrule{8-9}\cmidrule{11-12}
      &\bfseries Variable & \bfseries Factual && \bfseries Proto & \bfseries Proto + CEILS && \bfseries Proto & \bfseries Proto + CEILS && \bfseries Proto & \bfseries Proto + CEILS\\
      \midrule

      & $Y$ & 0 && 1  & 1 \\
      & \texttt{Mek} & 27.9 && 27.9 & 6.8 && 0     & -21.1  && 31.2 & 0\\
      & \texttt{PKC} & 10.3 && 7.0 & 3.4 && -3.3 & -6.9  && -3.3 &  -6.9\\
      & \texttt{PKA} & 319.0 && 1211.3 & 850.3 && 892.3     & 531.3  && 881.1 & 511.2\\
      & \texttt{Raf} & 47.4 && 47.4 & 35.6 && 0 & -11.8  && 18.0 &  0\\

      \bottomrule
    \end{tabular}
    }
\end{table}

\subsection{Discussion}
\label{sec:discussion}

After discussing separately the results of each experiment, we can summarize the overall findings as follows: 
\begin{itemize}
    \item CEILS provides, in general, counterfactual explanations farther in feature space with respect to the baseline method.
    \item CEILS is almost always more efficient in providing valid counterfactuals. This is more pronounced and noticeable when there are non-actionable constraints. In this case, the baseline method maybe not able to provide actual counterfactual explanations or they may be too far to be considered valid, whereas CEILS can indirectly act also on non-actionable features taking into account causal influences. 
    \item There are cases (e.g. Sachs dataset experiment) in which feasibility constraints and SCM compliance may result in a form of friction to find valid counterfactuals, as it could be expected.
    \item Comparing CEILS and its baseline (non-causal) method in terms of effort to reach the explanation, i.e. confronting their costs (distance in latent space for CEILS and distance in feature space for Proto) almost always results in a better performance for CEILS, again due to its ability in exploiting causal relationships.
    \item If we take into account the underlying SCM for both approaches, then the baseline method exhibits a very poor performance, and in particular it falls short of suggesting feasible actions to reach valid counterfactuals (as argued in section ~\ref{sec:action_feature_space}).
\end{itemize}

These findings are completely in line with the fact that CEILS effectively focuses on searching the \emph{nearest counterfactual in latent space}, thus it is optimized to find the less expensive set of actions with respect to an assumed SCM, thus guaranteeing a valid recourse.

% ================================================================================================= %
\section{Conclusions \label{sec:conclusions}}

Against the background of a flourishing literature on Explainable AI and in particular on counterfactual explanations, we have proposed a new approach --- Counterfactual Explanations as Interventions in Latent Space (CEILS) -- with a twofold goal, namely to take into account causality in generating counterfactual explanations and to employ them to provide feasible recommendations for recourse, while at the same time having the important characteristic of being a methodology easily adaptable on top of existing counterfactual generator engines. The experimental results clearly show that there are cases in which the baseline generator would recommend explanation completely unfeasible with respect to the underlying causal structure, while our approach --- on top of the same generator --- is able to provide more realistic and reachable counterfactual profiles, often with less effort. 

This is a first attempt in the direction of the ambitious target of providing the end user with realistic explanations \emph{and} feasible recommendations to gain the desired output in automatic decision making processes.

As for future work, we will tackle some limitations of our methodology and open challenges in the field of counterfactual explanations.
Firstly, it would be important to relax the assumption of having a complete and reliable causal graph, and allow for the possibility of having a causal-aware generator with an underlying \emph{partial} DAG (e.g. \cite{mahajan2019preserving} discuss this point in their proposal). Secondly, it would be valuable to find methods to relax the assumption of having a completely deterministic SCM in the form of an additive noise model~\eqref{eq:ANM} (e.g. \cite{karimi2020algorithmic2} take steps in this direction). Another assumption that we should address more properly is that of causal sufficiency, namely the fact the DAGs account for all the common causes of the observed variables, which is indeed a strong requirement and virtually impossible to be validated.     
Moreover, in our experiments we employ the methodology of~\cite{alibi} as a baseline generator for its remarkable characteristic of guiding the optimization process towards regions of the feature space that are close in distribution with respect to the observed data: we would like to analyze in detail how this entangles with our approach of applying the counterfactual generator in latent space rather than in feature space. 

Finally, it would be really useful to embed our proposed methodology in user-interaction tools and perform studies both to validate our method, and also to improve it by taking into account user feedback, possibly allowing the users to change, among other parameters, the feasibility constraints on actions.

\section*{Acknowledgements}
We want to thank Aisha Naseer (Fujitsu), Greta Greco (Intesa Sanpaolo) and Ilaria Penco (Intesa Sanpaolo) for their support in the joint collaboration that has lead to this work.

\section*{Disclaimer}
The views and opinions expressed within this paper are those of the authors and do not necessarily reflect the official policy or position of Intesa Sanpaolo or Fujitsu. Assumptions made in the analysis, assessments, methodologies, models and results are not reflective of the position of any entity other than the authors.

\singlespacing

\bibliographystyle{plainnat}    
\bibliography{ref.bib}   % name your BibTeX data base

\end{document}